\pdfoutput=1
\documentclass{article}

 \usepackage[preprint,nonatbib]{nips_2018_c}

\usepackage{array}

\usepackage{pdfpages}
\usepackage{algorithm}
\usepackage{algpseudocode}

\usepackage{graphicx}
\usepackage{amsmath}
\usepackage{relsize}
\usepackage{amssymb}

\usepackage{mathtools}
\usepackage{dsfont}
\usepackage{mathrsfs}

\usepackage{textcomp}
\usepackage{tikz}
\usetikzlibrary{positioning,calc}
\usetikzlibrary{fit}
\usetikzlibrary{shapes,arrows,shadows}

\makeatletter
\usepackage{wrapfig}
\usepackage{floatflt}
\usepackage{epstopdf}
\usepackage{caption}
\usepackage{float}
\usepackage{accents}
\usepackage[english]{babel}
\usepackage{listings}
\usepackage{color}
\usepackage{multirow}

\definecolor{mygreen}{rgb}{0,0.4,0}
\definecolor{mygray}{rgb}{0.5,0.5,0.5}
\definecolor{stringColor}{rgb}{0.88,0.1,0.12}
\usepackage{fancyvrb} 
\usepackage{comment,url}
\usepackage{enumerate}

\newcommand{\real}{\mathbb{R}}

\renewcommand{\tilde}[1]{\widetilde{#1}}

\def\app#1#2{%
	\mathrel{%
		\setbox0=\hbox{$#1\sim$}%
		\setbox2=\hbox{%
			\rlap{\hbox{$#1\propto$}}%
			\lower1.1\ht0\box0%
		}%
		\raise0.25\ht2\box2%
	}%
}

\definecolor{turr}{rgb}{0.4,0.97,0.97}

\usepackage[utf8]{inputenc} 
\usepackage[T1]{fontenc}     
\usepackage{hyperref}          
\usepackage{url}            	   
\usepackage{booktabs}        
\usepackage{amsfonts}        
\usepackage{nicefrac}          
\usepackage{microtype}       

\title{Latent Molecular Optimization for Targeted Therapeutic Design}

\author{
  Tristan Aumentado-Armstrong \\
  Department of Computer Science, 
  University of Toronto\\
  Vector Institute for Artificial Intelligence \\
  \texttt{tristan.aumentado@mail.utoronto.ca} 
}

\begin{document}

\maketitle

\begin{abstract}
We devise an approach for targeted molecular design, a problem of interest in computational drug discovery:
	given a target protein site, we wish to generate a chemical with both high binding affinity to the target and satisfactory pharmacological properties.
This problem is made difficult by the enormity and discreteness of the space of potential therapeutics, as well as the graph-structured nature of biomolecular surface sites.
Using a dataset of protein-ligand complexes, we surmount these issues by extracting a signature of the target site with a graph convolutional network and by encoding the discrete chemical into a continuous latent vector space.
The latter embedding permits gradient-based optimization in molecular space, which we perform using learned differentiable models of binding affinity and other pharmacological properties.  
We show 
that our approach is able to efficiently optimize these multiple objectives
and discover new molecules with potentially useful binding properties, validated via docking methods. 
\end{abstract}

\section{Introduction}
The goal of computational drug design is to assist in the discovery of molecules with therapeutic potential.
Given the vast and rapidly expanding knowledge from molecular biology, 
one may know the cellular machinery (usually a protein) responsible for a given disease,
and can thus devise a \textit{targeted}  therapeutic, designed to affect a particular biomolecule within the target pathway. 
This paradigm, known as molecular targeting therapy, has shown promise in therapeutic development (e.g.\ cancer \cite{tsuruo2003molecular,weinstein2006mechanisms}).
However, discovering and constructing a molecule (or ligand) capable of binding a given target is non-trivial.
The space of small molecules with therapeutic potential is estimated to be on the order of $10^{60}$ \cite{bohacek1996art,reymond2012enumeration}.
More problematic is the discreteness of chemical structure space, in which minor perturbations can lead to large changes in biochemical properties (such as affinity), preventing continuous optimization approaches.
Further, strong binding affinity to the target is merely necessary, not sufficient, for a useful therapeutic:
the presence of off-target effects, cost, and pharmacological
ADMET (absorption, distribution, metabolism, excretion, and toxicity) properties
also play a role in the efficacy or viability of a drug \cite{paul2010improve,leeson2007influence}.
Therapeutic design is therefore a multi-objective optimization problem of considerable difficulty,
causing the pharmaceutical industry to struggle to find viable ligands 
able to reach the end of the development pipeline
\cite{scannell2012diagnosing,dimasi2016innovation}.

One promising route to alleviating this problem is the use of computational methods, 
such as virtual screening \cite{tanrikulu2013holistic}, molecular docking \cite{ferreira2015molecular}, and quantitative structure-activity relationship (QSAR) models \cite{dudek2006computational,verma20103d}, all of which attempt to predict or characterize affinity of a small molecule to a target.
More recently, the advent of deep learning has improved various prediction tasks in cheminformatics \cite{gawehn2016deep,chen2018rise}, such as
bioactivity scoring \cite{wallach2015atomnet,ragoza2017protein}, organic synthesis planning \cite{segler2018planning}, and toxicity estimation \cite{unterthiner2015toxicity}.
Of particular interest, however, is the \textit{de novo} design of molecules with desirable properties \cite{benhenda2017chemgan,schneider2016novo} (also known as the inverse QSAR problem  \cite{wong2009constructive,miyao2010exhaustive,miyao2016inverse}),
which has been approached in a variety of ways, including
Bayesian optimization in latent space \cite{gomez2016automatic,daisyntax,jin2018junction,kusner2017grammar,griffiths2017constrained},
reinforcement learning \cite{olivecrona2017molecular},
deep generative models \cite{sanchez2017optimizing,kadurin2017drugan}, 
and
recurrent neural networks (RNNs) \cite{jaques2016sequence,segler2017generating}.
Many of these methods, however, do not address the problem of targeted design, while those that do suffer from short-comings, such as relying on specialized biochemical models, requiring bound complexes for affinity estimation, or not utilizing the structural data contained in other protein-ligand pairs.

In this work, we present a novel method to perform targeted design of small molecule agents via optimization in the latent space of a generative model. 
First, using a dataset of protein-ligand complexes (PLCs), we extract and separate 
the protein binding sites as graph-valued data and the ligands as discrete chemical SMILES strings \cite{anderson1987smiles}.
A graph convolutional approach is used to map the protein sites to a vector signature $P$, 
	while the Junction Tree Variational Autoencoder (JTVAE) \cite{jin2018junction} is used to map the discrete chemical to a latent vector $C$.
Several differentiable models are then learned via neural networks:
(1) a direct mapper, 
which attempts to reconstruct $C$ from $P$ alone,
(2) a binding affinity estimator, 
and
(3) intrinsic property regressors, which predict the ease of synthesis, drug-likeness, and toxicity of $C$.
This pipeline is illustrated in Fig.\ \ref{desc}.
All these models are utilized in a gradient-based optimization algorithm over latent chemical space, which is given only an embedded target protein site $P$ and is tasked with finding a small molecule with both strong binding affinity to the target as well as desirable pharmacological properties.

Our main contributions are to provide ways to circumvent some of the limitations of current approaches.
First, by working in a latent vector space, we can use continuous optimization techniques to perform efficient gradient-based search over chemical structures.
Second, by utilizing a generic representation of protein surface sites, we can take advantage of the information held in datasets of bound protein-ligand pairs, meaning we do not necessarily require large amounts of biochemical data for a target protein (i.e.\ we attempt to generalize properties of other complexes to our target).
Third, we avoid docking by learning a differentiable scoring function that takes in a protein surface site and ligand separately.
To our knowledge, this is the first work utilizing these techniques for targeted molecular design, which we hope will prove useful for future endeavors in this area.

\tikzset{%
	block/.style    = {draw, thick, rectangle, minimum height = 1.9em,
		minimum width = 1.9em},
	fm/.style      = {draw, circle, node distance = 1.7cm}, 
	fm2/.style      = {draw, circle, node distance = 1.7cm}, 
	input/.style    = {coordinate}, 
	output/.style   = {coordinate}, 
	fitting node/.style={
		inner sep=0pt,
		fill=none,
		draw=none,
		reset transform,
		fit={(\pgf@pathminx,\pgf@pathminy) (\pgf@pathmaxx,\pgf@pathmaxy)}
	},
	reset transform/.code={\pgftransformreset}
}
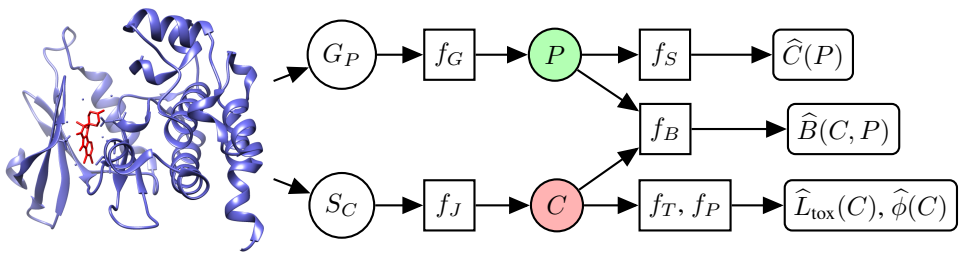
\begin{figure}
	\centering
	\begin{tikzpicture}[auto, thick, node distance=1.9cm, >=triangle 45] 
	\draw 
	node at (0,0)[name=pic] {\hspace*{-4mm}\includegraphics[width=0.25\textwidth]{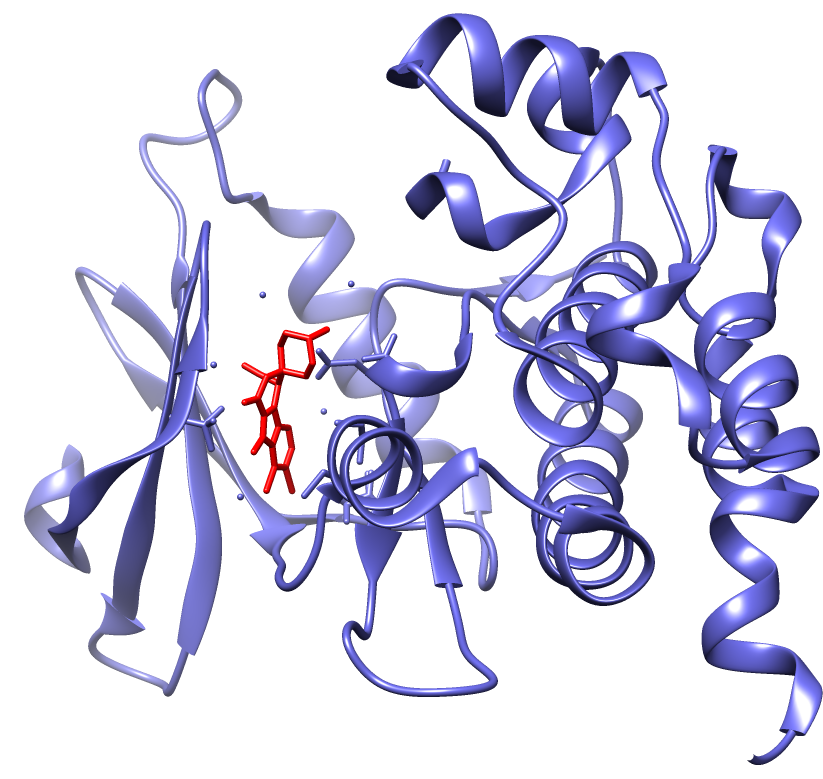}}
	node [fm,above right of=pic,xshift=13.977mm,yshift=-2mm] (gp) {$ G_P$}
	node [fm,below right of=pic,xshift=13.977mm,yshift=2mm] (sc) {$ S_C$}
	node [block,right of=gp,xshift=-4.7mm,yshift=0mm] (fg) {$ f_G$}
	node [block,right of=sc,xshift=-4.7mm,yshift=0mm] (fj) {$ f_J$}
	node [fm,right of=fg,name=p,fill=green!30,xshift=-2.97mm]{$P$}
	node [fm,right of=fj,fill=red!30,node distance=1.95cm,xshift=-5.197mm] (c) {$C$}
	node [block,right of=p,xshift=-4.37mm] (fs) {$ f_S$}
	node [block,above right of=c,yshift=-3.279mm,xshift=0.97mm] (fb) {$ f_B$}
	node [block,right of=c,xshift=-1.97mm] (ft) {$ f_T$, $f_P$}
	node [block,rounded corners=0.1cm,right of=fs,xshift=0.7mm] (os) {$ \widehat{C}(P)$}
	node [block,rounded corners=0.1cm,right of=fb,xshift=4.7mm] (ob) {$\widehat{B}(C,P)$}
	node [block,rounded corners=0.1cm,right of=ft,xshift=5.7mm] (ot) {$ \widehat{L}_\text{tox}(C)$, $ \widehat{\phi}(C)$};
	\draw[->](p) -- (fs);
	\draw[->](p) -- (fb);
	\draw[->](c) -- (fb);
	\draw[->](c) -- (ft);		
	\draw[->](ft) -- (ot);
	\draw[->](fs) -- (os);
	\draw[->](fg) -- (p);
	\draw[->](fj) -- (c);
	\draw[->](gp) -- (fg);
	\draw[->](sc) -- (fj);
	\draw[->](pic) -- (gp);
	\draw[->](pic) -- (sc);
	\draw[->](fb) -- (ob);
	\end{tikzpicture}
	\caption{Schematic overview of model components. 
		Graph site $G_P$ and chemical ligand structure $S_C$ are extracted from an input PLC (PDB: 3BHY), and embedded into a vector space via $f_G$ and $f_J$. The output vectors $P$ and $C$ are then used to compute a predicted binder $ f_S(P) = \widehat{C}(P)$, estimated binding affinity $ f_B(C,P) = \widehat{B}(C,P) = (p_B, \widehat{B}_{\text{DSX}})$, and predicted chemical properties 
		(toxicity $f_T(C) = \widehat{L}_\text{tox}(C) $ and
		drug-likeness $f_P(C) = \widehat{\phi}(C)$). 
	}
	\label{desc}\end{figure}

\section{Related Work} 

Given the importance of \textit{de novo} chemical design to the pharmaceutical development process,
	a considerable body of work has been devoted to computational methods for improving it 
	(e.g.\ \cite{chen2018rise,schneider2016novo}).
Much recent research on constructing continuous embeddings of chemicals has used Bayesian optimization to generate molecules with desirable properties \cite{gomez2016automatic,daisyntax,jin2018junction,kusner2017grammar,jaques2016sequence,griffiths2017constrained}. 
However, these methods do not perform \textit{targeted} chemical design, aimed at a particular molecule of interest. 

Inspired by the success of stochastic gradient-based optimization in training deep neural networks, we also take advantage of the continuous nature of the latent chemical space, using ADAM \cite{kingma2014adam} for optimization, rather than a Bayesian approach. 
This has the disadvantage of requiring differentiable models for all of the quantities we wish to optimize; however, deep learning models can be used to satisfy this requirement.

Separately, methods of predicting affinity from structural information have also been recently developed \cite{ragoza2017protein,wallach2015atomnet}. 
Such networks take a PLC as input, and use the structural information from the complex to score the affinity of the bound pair.
The main difference between our method and these approaches is that our affinity prediction network does not receive structural information on the interaction; i.e.\ the protein site and putative binding chemical are received independently, as embedded vectors. 
This makes the estimation problem more difficult, since detailed atomic interaction information is not available to the predictor, but more general, since it can be applied to any protein site-chemical pair.
This allows us to avoid a docking step (i.e.\ to generate a PLC) during our optimization.

The most similar studies to ours perform targeted molecular generation.
For instance, the work by Takeda et al.\ \cite{takeda2016chemical}
	assists \textit{de novo} design by visualization of various properties over chemical space.
Other examples include works using RNNs \cite{olivecrona2017molecular,segler2017generating} to generate SMILES strings with desired properties, in particular affinity to a target; these works, as well as the work by Blachke et al.\ \cite{blaschke2017application}, train their models on biological activity data for specific protein targets (i.e.\ using a specific QSAR model). 
In other words, these models will work best on target protein molecules that already have considerable data available, concerning known binders and their affinities.
This is in contrast to our method, which attempts to perform a generic mapping from \textit{any} given target protein site (i.e.\ ``patch'' of atoms from a protein) to a binding chemical.
As such, our approach is to rely on learning a general representation of affinity, based on structural information, rather than specializing to a specific input.
Thus, our algorithm can be applied to targets for which no biochemical data is known; the predictions of the model will be based on the patterns learned from \textit{other} complexes.
The price to pay for this generality is the need for structural information; however, the exponential growth of such data bodes well for this paradigm \cite{burley2017protein}.
Of course, when such activity data is available for a target, it should be utilized; 
such information could easily be integrated into our algorithm, simply by adding another term to the energy function, based on the activity predicted from the specific QSAR model.

\section{Molecular Representations}

\subsection{Latent Chemical Embedding}

Considerable progress has been made using deep generative models to embed discrete chemical structures into continuous latent vector spaces \cite{gomez2016automatic,daisyntax,kusner2017grammar,janz2017learning}.
Such spaces allow the use of continuous optimization approaches in the latent space to perform efficient search over chemical structures.
In this work, we use the JTVAE via a pretrained model with latent dimensionality 56 (see \cite{jin2018junction} for details).
Briefly, this method improves over previous SMILES-based methods by working directly in the space of graphs, 
resulting in a higher probability of decoding a valid molecule (i.e.\ one is less likely to encounter areas of latent space that do not correspond to valid molecules).
It utilizes a two-part molecular encoding: 
	a junction tree encoding, 
		which contains information on substructures within the chemical,
	and
	a graph encoding, 
		which holds details on the atomic connectivity.
We use the concatenation of these two vectors as our latent representation.
RDKit \cite{landrum2006rdkit} was used for chemical structure processing.
Throughout the paper, we denote $\mathcal{L}$ as the latent space of chemicals. 
The JTVAE thus induces a map $f_J : \mathscr{S}_C \rightarrow \mathcal{L}$. 
We write $S_C  \in \mathscr{S}_C$ and $C \in \mathcal{L}$ to mean a discrete chemical structure and its latent form, respectively (as in Fig.\ \ref{desc}).
 
\subsection{Protein Site Signatures via Graph Convolutions}

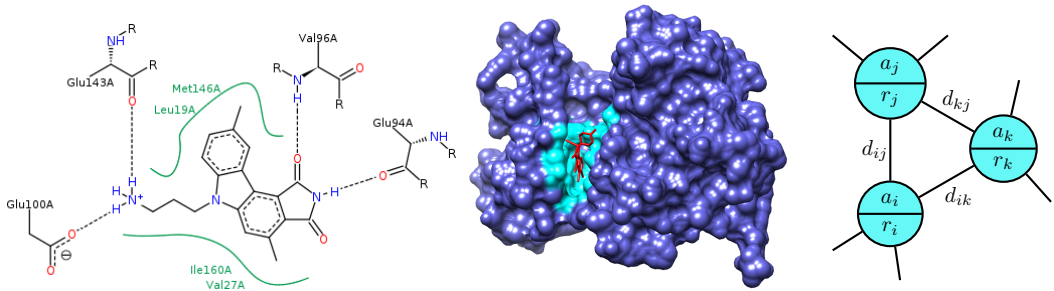
\begin{figure}
	\centering
	\includegraphics[width=0.42735\textwidth]{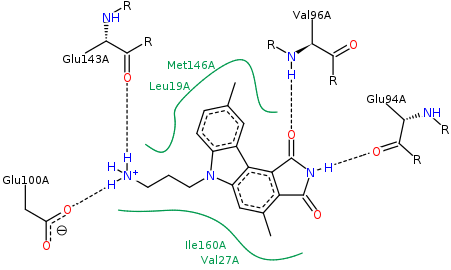}\hfill
	\includegraphics[width=0.320\textwidth]{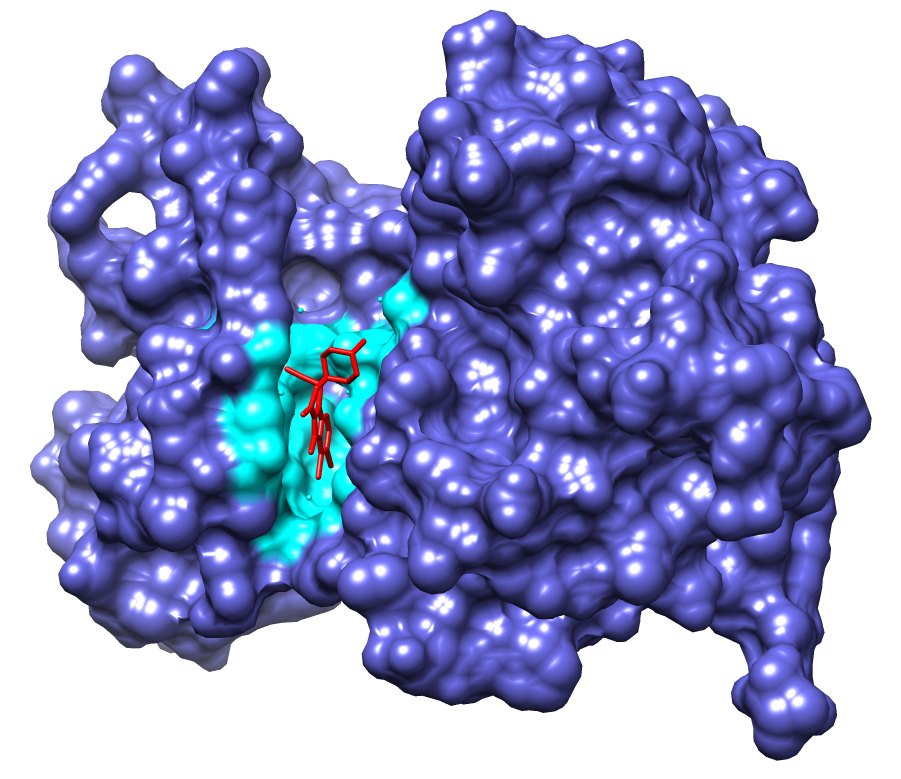}\hfill 
\tikzstyle{abstract}=[circle, draw=black, fill=white]
\tikzstyle{labelnode}=[circle, draw=white, fill=white]
\tikzstyle{line} = [draw, -latex']
	\begin{tikzpicture}[
	qq/.style={circle split, draw, minimum size=0.1cm, fill=turr},
	scale=0.82, every node/.style={scale=0.82}
	]
	\node (ii) at (10bp,29bp) [qq,thick] {$a_i$ \nodepart{lower} $r_i$};
	\node (jj) at (10bp,89bp) [qq,thick] {$a_j$ \nodepart{lower} $r_j$};
	\node (kk) at (62bp,59bp) [qq,thick] {$a_k$ \nodepart{lower} $r_k$};
	\node (n1i) at (15bp,-5bp) [] {};
	\node (n2i) at (-20bp,10bp) [] {};
	\node (n1j) at (40bp,114bp) [] {};
	\node (n2j) at (-20bp,114bp) [] {};
	\node (n1k) at (70bp,94bp) [] {};
	\node (n2k) at (87bp,31bp) [] {};
	\draw [thick] (ii) -- (jj) node [midway, left, xshift=1.0mm] {$d_{ij}$}; 
	\draw [thick] (ii) -- (kk) node [midway, below, xshift=2.0mm, yshift=0.5mm] {$d_{ik}$}; 
	\draw [thick] (kk) -- (jj) node [midway, above, xshift=1.5mm, yshift=-0.5mm] {$d_{kj}$}; 
	\draw [thick] (ii) -- (n1i);
	\draw [thick] (ii) -- (n2i);
	\draw [thick] (jj) -- (n1j);
	\draw [thick] (jj) -- (n2j);
	\draw [thick] (kk) -- (n1k);
	\draw [thick] (kk) -- (n2k);
	\end{tikzpicture}
	\caption{Protein binding and target site characterization. 
		Left: 2D Poseview \cite{stierand2010drawing} diagram of chemical interactions between ligand and protein residues (PDB: 1WVX), where dashed lines mark hydrogen bonds and green curves demarcate the protein surface. 
		Middle: 3D depiction in Chimera \cite{pettersen2004ucsf} of a protein-ligand complex (PDB: 3BHY) with surface filled in (red: chemical ligand, blue: protein, turqoise: binding site). 
		Right: graph-theoretic representation of binding site (nodes are atoms with atomic and residue identity as features; edge features are based on inter-atomic distances).}	
	\label{graph}
\end{figure}

Inspired by the use of graph convolutional networks (GCNs) in extracting fingerprints from chemical structure graphs \cite{duvenaud2015convolutional} and processing protein interface sites \cite{fout2017protein}, we chose to represent our target site in a graph-theoretic manner (see Fig.\ \ref{graph}, right inset).
Each site consists of the nitrogen, oxygen, and carbon atoms within 4\AA\ of the bound ligand as the graph vertices, with features given by the concatenation of two one-hot vectors, each of length 24: 
(1) the atomic identity, including the remoteness indicator, and (2) the amino acid residue identity to which the atom belongs (including four non-standard residues).
The graph is then treated as fully connected, 
with edge weights given by:
$ e_{ij} = 1 / (1 + d_{ij}^2) $, where $d_{ij}$ is the Euclidean distance between nodes (atoms) $i$ and $j$.
Protein structure processing was done with Biopython \cite{doi:10.1093/bioinformatics/btp163}, DeepChem \cite{deepchem}, and MDTraj \cite{McGibbon2015MDTraj}.

Our GCN is based on a combination of the fingerprint extraction approach of Duvenaud et al.\ \cite{duvenaud2015convolutional} and the approximate convolution method of Kipf and Welling \cite{kipf2016semi}. 
Consider a single input graph $G_P$ for a protein site, with $N_a$ atoms and node feature dimensionality $N_F$.
Define the space of extracted signatures to be a vector space $\mathcal{P}$ with dimensionality $N_P$; note that $N_P$ is a hyper-parameter, while $N_a$ varies for each input protein.
Denote $A\in \real^{N_a \times N_a}$ as the weighted adjacency matrix (i.e.\ with entries $e_{ij}$), 
$D_{ii} = \sum_j A_{ij}$, and $V\in \real^{N_a\times N_F}$ as the matrix of node feature vectors. 
Let $\sigma_p$ denote a point-wise non-linearity  and $ g_{\text{SM} } $ denote the row-wise softmax function.
Then, for every layer $\ell$ in the GCN, we have two matrices of trainable weights:
	$ W^{(\ell)}\in\real^{N_F\times N_F} $, which controls the convolutional parameters,
	and
	$ \tilde{W}^{(\ell)}\in\real^{N_F\times N_P} $, which determines the signature extraction. 
Thus, given inputs $V$, $D$, and $A$, we can compute
\begin{equation}
H^{(\ell+1)} = \sigma_p(D^{-1/2} A D^{-1/2} H^{(\ell)} W^{(\ell)} )
\end{equation}
where $ V = H^{(0)} $. The final vector $P \in\mathcal{P}$ is then given by
\begin{equation}
	P = \sum_{\ell = 1}^{L} \sum_{k=1}^{N_a}  g_{ \text{SM} }( H^{(\ell)} \tilde{W}^{(\ell)} )_k
\end{equation}
where $L$ is the total number of layers  and $g_{\text{SM} }(M)_k$ refers to the $k$th row of the output matrix. 
Here, we used $L=2$, $\sigma_p = $ ReLU, and $N_P = 100$.
We can thus denote our GCN as a map performing $ f_G( G_P ) = P $, as in Fig.\ \ref{desc}.

\section{Binding Affinity Regression and Direct Mapping}

\subsection{Affinity Data}
The core mechanism of our targeting algorithm relies on a differentiable predictor of binding affinity.
We therefore required a dataset of protein-ligand complexes from which to learn (see Fig.\ \ref{graph} for visualization of a PLC and binding site). 
We chose to use the scPDB \cite{kellenberger2006sc,meslamani2011sc,desaphy2014sc}, a database of protein-ligand complexes from the PDB \cite{rose2016rcsb} amenable to cheminformatics studies. 
Such complexes are sufficient to estimate a probability of binding, by classifying whether or not a given protein site and chemical interact.
We also examined the presence of affinity data with structural counterparts in the BMOAD database \cite{ahmed2014recent}. 
This data is useful for evaluating the quality of binding, e.g.\ differentiating between the level of affinity for two chemicals binding the same protein.

However, due to the heterogeneity and reduced dataset coverage of the empirical binding affinity data, we chose to instead use a computational scoring function, which is an \textit{in silico} method of estimating binding quality, used for tasks like molecular docking \cite{huang2010scoring}.
Scoring functions are not necessarily well-correlated to experimentally determined affinities \cite{kastritis2010scoring,grinter2014challenges}, though ideally this is a preferable property.
We therefore chose to use DrugScoreX (DSX) \cite{neudert2011dsx}, as prior studies had found decent correlation to empirical affinities \cite{cao2014improved}, a scoring function based on the older DrugScore and DrugScore$^\text{CSD}$ functions \cite{gohlke2000knowledge,velec2005drugscorecsd}.
DSX is a ``knowledge-based potential'', 
using statistical analysis of atomic interactions to construct a distance-dependent potential able to assess binding affinity \cite{liu2015classification}.
We ran DSX on all the PLCs in the scPDB to obtain an affinity score for each bound complex.
To verify correlation to empirical binding data, we compared the DSX results to the experimental results in BMOAD, finding significant positive correlations between DSX score and the various experimental affinities
(see Supplementary Material). 
Our final dataset was separated into a training/validation set of size 16090, and a held-out test set of size 1000.

\subsection{Models}

\begin{figure}
	\includegraphics[width=0.32930129\textwidth,trim={9mm 9mm 9mm 9mm},clip]{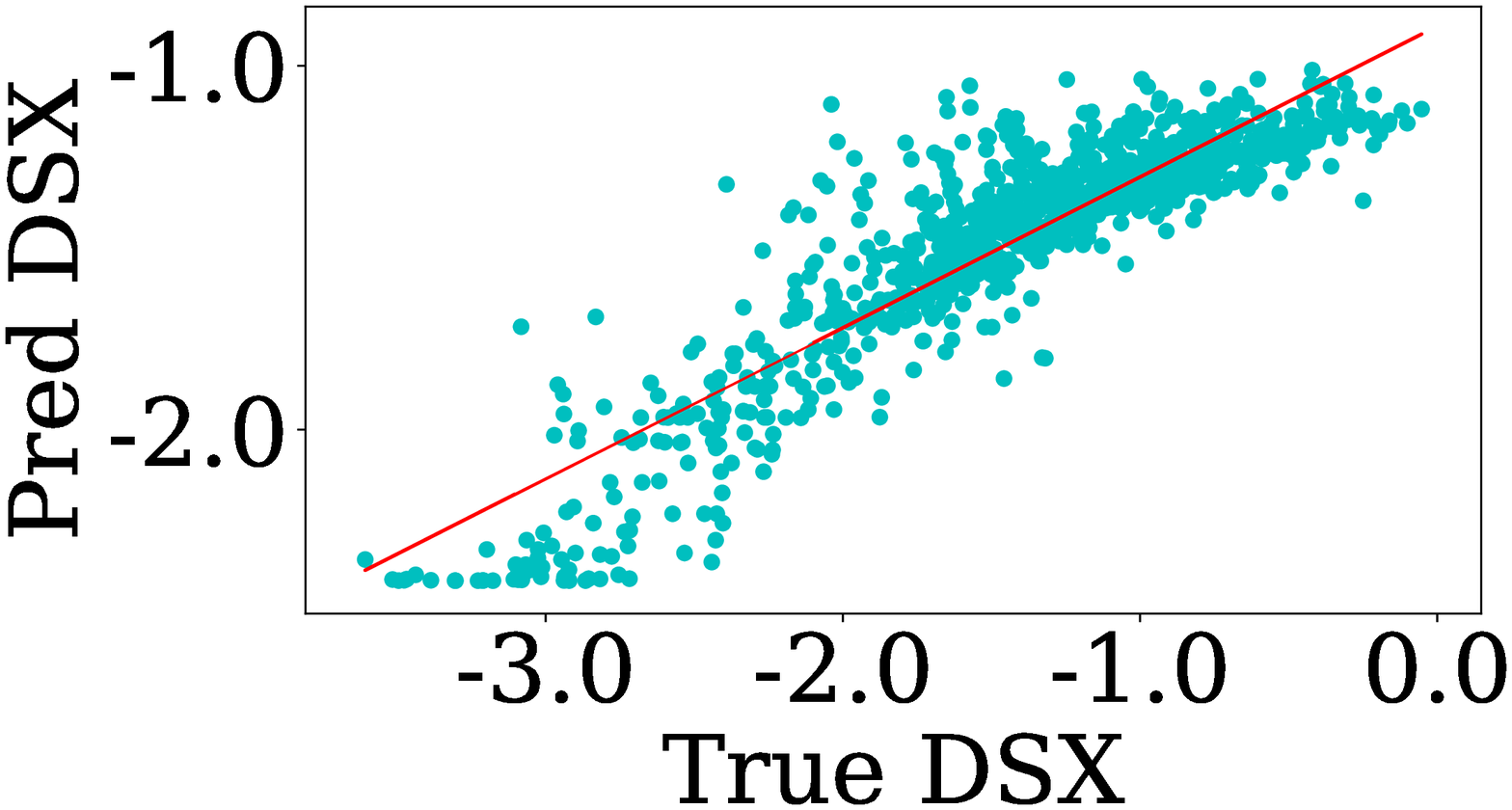} \hfill 
	\includegraphics[width=0.32930129\textwidth,trim={9mm 9mm 9mm 9mm},clip]{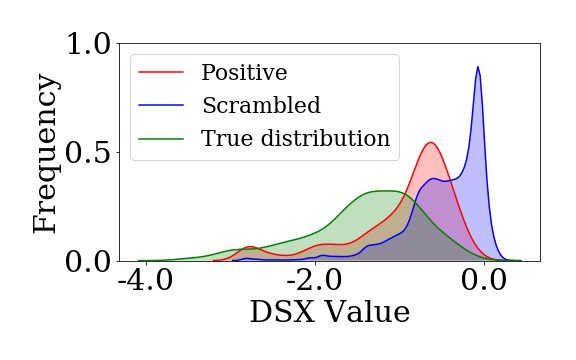} \hfill
	\includegraphics[width=0.32930129\textwidth,trim={9mm 9mm 9mm 9mm},clip]{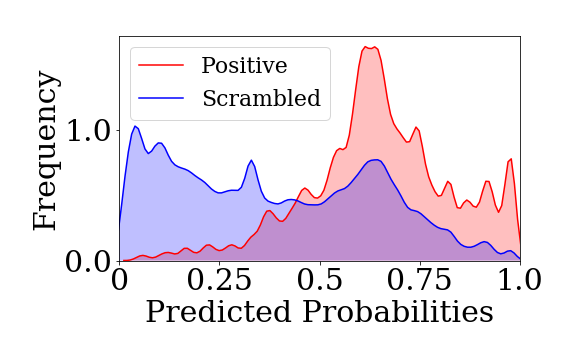} 
	\caption{
		Left: correlation plot between true and predicted DSX on positive data examples (in hundreds); line of best fit (red) has slope 0.82.
		Middle: normalized frequency plots of the predicted DSX (in hundreds) on scrambled examples (blue), positive examples (red), and the true DSX distribution on positive examples (green).
		Right: normalized frequency plots of the binding probabilities on the positive (red) and negative (blue) examples. }
	\label{affinityres}
\end{figure}

Denote the set of PLCs as a set of tuples $(C,P)$, representing a latent chemical and embedded protein site produced by $f_J$ and $f_G$ respectively, where each tuple has an associated binding affinity $ {B}_{\text{DSX}}\in\real $.
We learned two models from this data:
(1) $f_B : \mathcal{L} \times \mathcal{P} \rightarrow \real^2 $
and
(2) $ f_S : \mathcal{P} \rightarrow \mathcal{L} $.
The first model is our affinity model 
 $ f_B(C,P) = \widehat{B}(C,P) = (p_B, \widehat{B}_{\text{DSX}}) $, 
 where $ p_B $ is the probability of binding and $\widehat{B}_{\text{DSX}}$ is the predicted DSX (i.e.\ estimated quality of binding).
The second model can be called a ``direct mapper'': given the input protein site $P$, it attempts to directly construct a binding ligand structure $C$.
Given the size of chemical space and the potential multiplicity of binders per target site, we do not expect $f_S$ to exactly reproduce the partner ligand (else it may be overfitting); rather, we simply hope that it maps $P$ to an area in latent space that is favourable to binding $P$. 
The direct mapper can therefore be used to initialize our gradient descent optimization algorithm.

The affinity model was implemented as a neural network with input batch normalization and two hidden layers (sizes 100 and 50), and regularized with dropout ($p=0.25$) and $L_2$ weight decay.
The output $p_B$ was also run through a sigmoid function and utilized binary cross-entropy loss, while $\widehat{B}_{\text{DSX}}$ used mean-squared error (MSE) loss.
The direct mapper had the same architecture, but with layer sizes 150 and 75, dropout probability $p=0.4$, and MSE loss. 
All networks in this paper were implemented in PyTorch, used ReLU, and were trained with ADAM \cite{kingma2014adam}.

We jointly trained the models $f_G$ (the GCN that outputs $P$), $f_B$, and $f_S$  (see Fig.\ \ref{desc}). 
Note that the losses of both $f_B$ and $f_S$ affect $f_G$ (i.e.\ both shape the embedding space defined by the protein site vector signature extractor).
To train the predictor, we used the known PLCs as positive examples; however, for negative examples we ``scrambled'' the tuples (i.e.\ matched a random $C$ to a $P$ with which it did not interact).
Negative examples were given a DSX of zero (note that more negative DSX means better binding).
Every minibatch was half positive and half negative examples.

We show the results of the affinity model on the held-out test set in Fig.\ \ref{affinityres}.
The results on scrambled data are computed via 100 random scrambled samples per positive example (thus there are 100 times more negative examples).
The Pearson and Spearman correlation (PC/SC) between the true and predicted DSX are 0.9 and 0.85 respectively 
(with significance $p < 0.01$), showing good agreement between the true DSX and the output of the regressor on positive examples. The DSX predictor does have more difficulty differentiating scrambled from unscrambled data, judging from the overlap between the blue and red distributions (Fig.\ \ref{affinityres}, middle inset); however, the positive distribution is clearly much closer to the true distribution, 
and there is a clear spike near zero for the scrambled DSX values.
The mean values with standard error (SE) are $-140.7 \pm 2.1 $, $ -94.9 \pm 2.0 $, and $ -46.1 \pm 0.14 $ for the true data, positive predictions, and scrambled predictions, respectively.
For the binding probabilities, the average probability (with SE) for a positive pair was 
$0.651 \pm 0.006$, while the average probability for a scrambled pair was $0.38 \pm 0.001$ (AUROC: 0.79).

Evaluating the direct mapper $f_S$ is more difficult. As noted above, we cannot expect it to output the exact answer; furthermore, simply reporting the Euclidean distance in latent space is a highly unintuitive measure of error. 
We therefore define a metric $R_E(P,C)$, 
which is designed to measure how much closer the output molecule 
$\widehat{C} = f_S(P)$ is to the true binder agent $C$,
compared to a \textit{randomly chosen} chemical $\tilde{C}$ from the prior over our latent space. 
We can thus define an error given by the \textit{log ratio of expected squared distance} as
\begin{equation}
	R_E(P,C) = \log\left( \mathbb{E}_{\tilde{C}\sim \mathcal{N}(0,I)}\left[ 
	\frac{|| C - \tilde{C} ||_2^2}{|| C - f_S(P) ||_2^2} \right] \right)
	= \log\left( \frac{||C||^2_2 + \text{dim}(\mathcal{L})}{|| C - f_S(P) ||_2^2} \right)
\end{equation}
which follows from $ || C - \tilde{C} ||_2^2 $ being non-central Chi-squared distributed.
When $R_E$ is equal to zero, then the direct mapper is at a distance no better than what we expect to attain by random chance; when $R_E < 0$, the direct mapper is doing worse than randomly sampling from the prior. Thus, we hope to see a positive $R_E$ score, which implies $f_S$ maps $P$ to a part of $\mathcal{L}$ that is closer to $C$ than random. 
We assume that the neighbourhood of a binder is more likely to favour producing other binders of $P$.

The results of the direct mapper are shown in Fig.\ \ref{chemprop} (bottom row, middle and right).
The vast majority of the data are greater than zero, indicating that the direct mapper is able to map closer to the true binder than one would expect by chance.
We show the results on both the training and testing set: 
	means with SE are $ 0.467 \pm 0.01 $ and $ 0.434 \pm 0.004 $.
Given their similarity, we are confident that the model is not overfitting. 
As a side note, it is possible that $\mathcal{L}$ could contain the binders of $P$ in disjoint areas of space, which would be problematic for the direct mapper $f_S$. 
This could potentially be mitigated by jointly training $f_J$ as well, but we leave this approach to future work.

\section{Chemical Property Prediction}

\begin{figure}
	\includegraphics[width=0.329\textwidth,trim={9mm 9mm 9mm 9mm},clip]{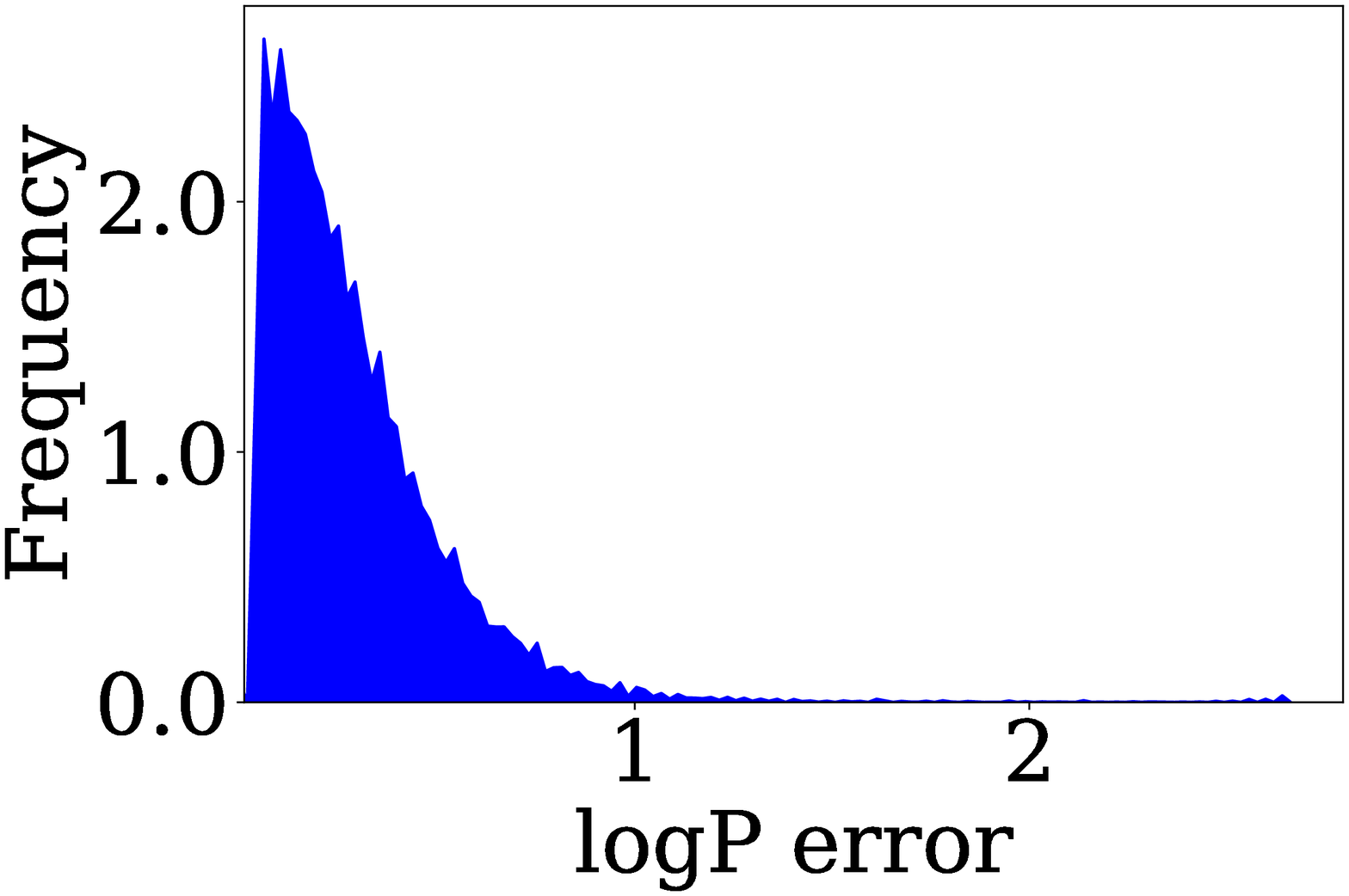} \hfill 
	\includegraphics[width=0.329\textwidth,trim={9mm 9mm 9mm 9mm},clip]{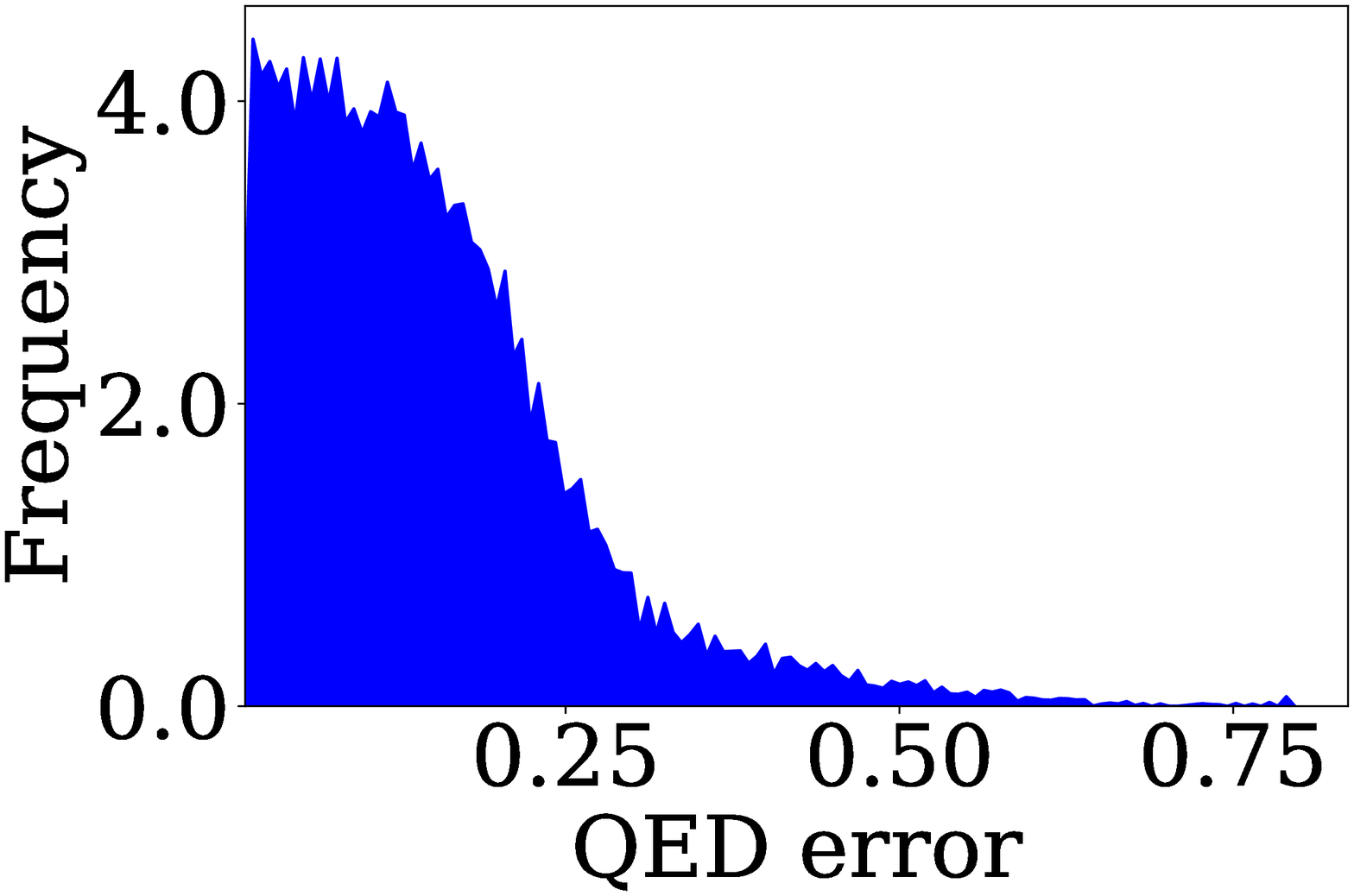} \hfill
	\includegraphics[width=0.329\textwidth,trim={9mm 9mm 9mm 9mm},clip]{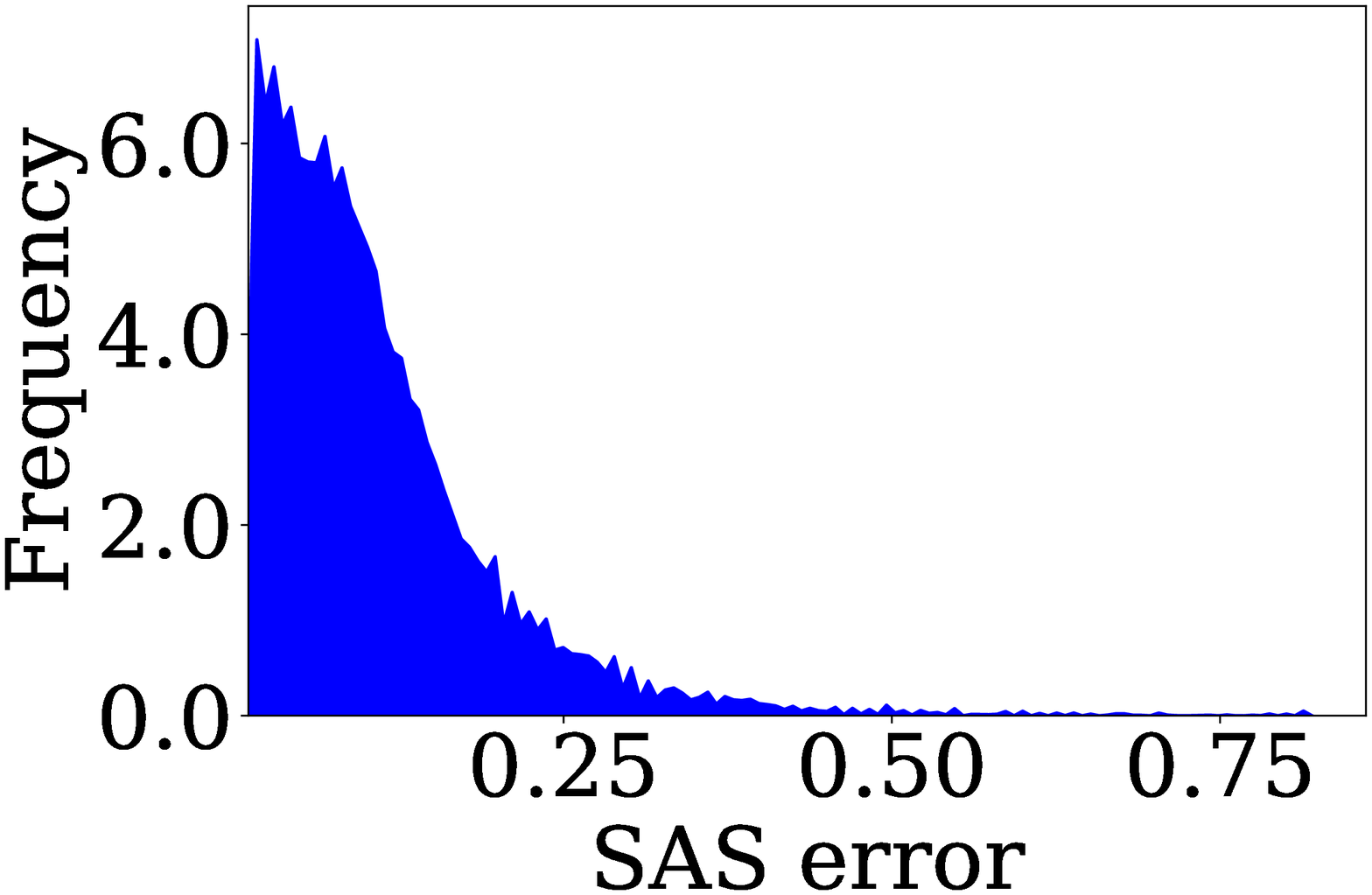} \\
	\includegraphics[width=0.329\textwidth,trim={9mm 9mm 9mm 9mm},clip]{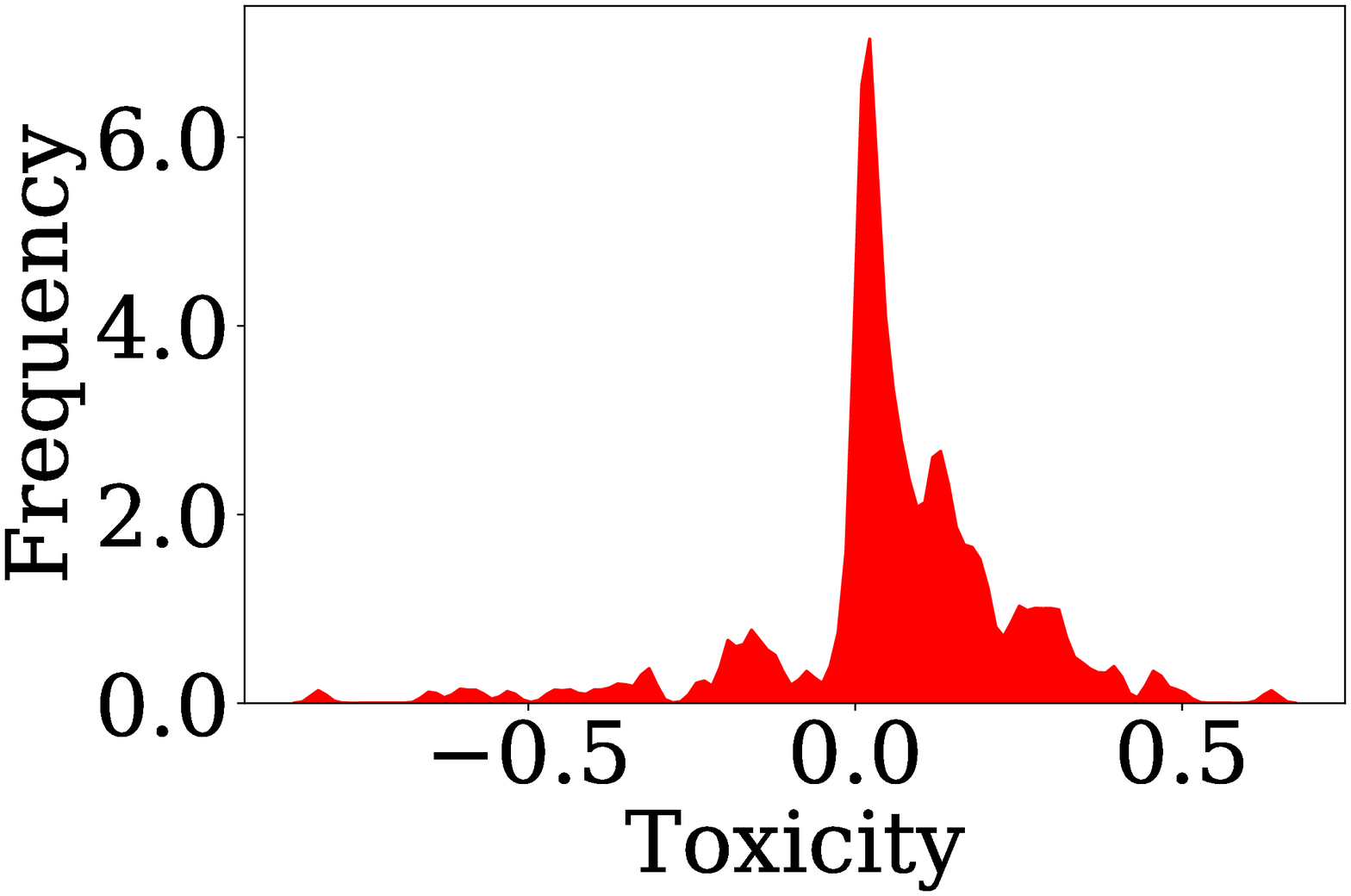} \hfill 
	\includegraphics[width=0.329\textwidth,trim={9mm 9mm 9mm 9mm},clip]{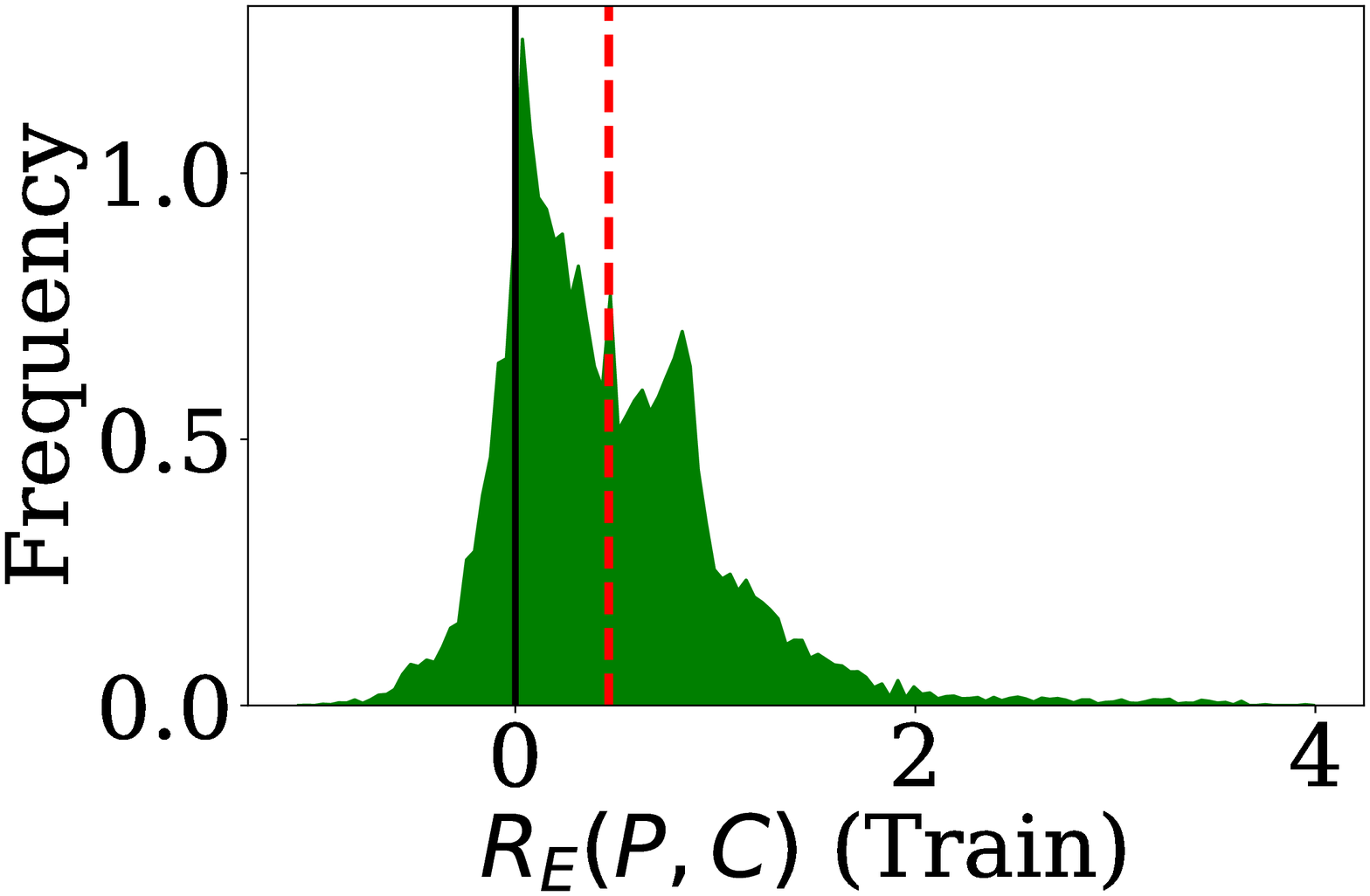} \hfill
	\includegraphics[width=0.329\textwidth,trim={9mm 9mm 9mm 9mm},clip]{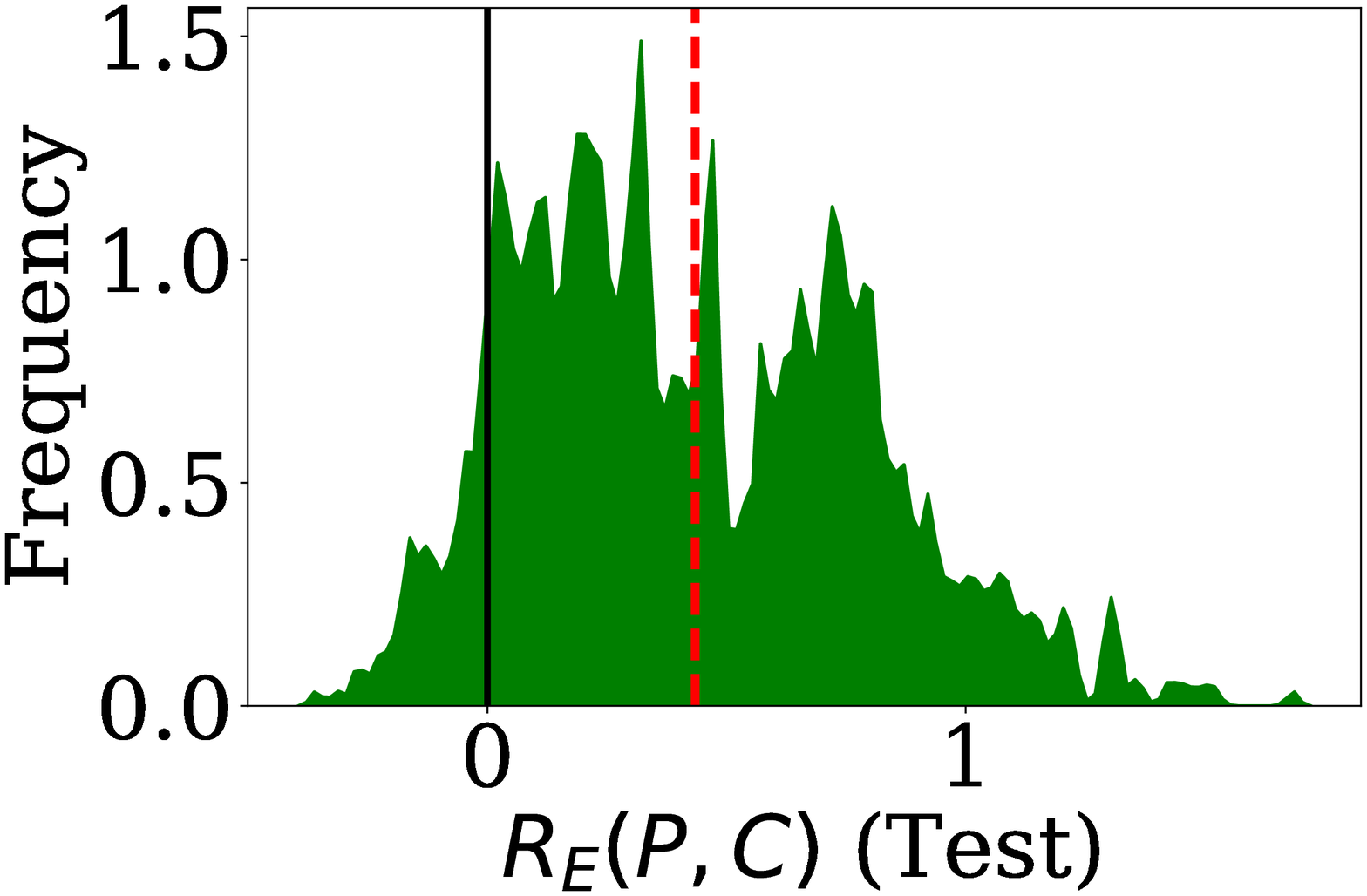} 
	\caption{Chemical property prediction and direct mapper results. Upper Row: relative error distribution on logP, QED, and SAS, respectively. Lower Row: error in toxicity prediction (left) and log ratio of expected squared distance error for the direct mapper (middle on the training set and right on the test set; the dashed red lines indicate the means of the $R_E$ values, at $ 0.467 \pm 0.01 $ and $ 0.434 \pm 0.004 $ with SE, for the training and testing data, respectively). }
	\label{chemprop}
\end{figure}

Targeted drug design cannot only take predicted binding strength into account; 
it must also consider molecular properties of a candidate therapeutic.
Indeed, molecules with good binding capabilities, but poor pharmacokinetics and toxicity properties, can result in costly failures as drugs \cite{waring2015analysis,kola2004can}.
We therefore consider the predicted toxicity and several drug-likeness properties of the chemical as well.

First, we considered estimating the toxicological properties of a given small molecule. 
We used the Tox21 dataset \cite{huang2016tox21challenge}, which includes binary data on the activation of five cellular stress response pathways. 
Not all chemicals have data for all toxicity measures; 
	hence, we use the average toxicity of the measures that were present as the output label, 
	denoted $ L_\text{tox}(C) $.
However, the dataset is very imbalanced,  
since most of the molecules have a zero average toxicity (see Supplementary Material).
We therefore linearly upweighted positive data points via:
$ W_\text{tox}(C) = (W_m-1) L_\text{tox}(C) + 1$,
where we fixed $W_m = 5 $ based on the frequency of positive values in the dataset.
For our estimator, let $ f_T : \mathcal{L} \rightarrow [0,1] $ be the toxicity predictor.
Its architecture is the same as that used to compute $p_B$, but without batch normalization.
Fig.\ \ref{chemprop} (bottom left) shows the distribution of the error 
$ \text{Err}_T = f_T(C) - L_\text{tox}(C) $ on a held-out test set of size 250 
(mean with SE of $|\text{Err}_T|$: $0.14 \pm 0.009 $; PC: $0.54$).
Note that the prediction errors are skewed to the positive values, due to the weighting function.

We also looked at three chemical properties, known to be associated with drug-likeness:
the partition coefficient logP, 
	which controls the lipophilicity (and thus affects the pharmacokinetic and ADMET properties of the molecule \cite{leeson2007influence,waring2010lipophilicity}),
the Quantitative Estimation of Drug-Likeness (QED), 
	which attempts to quantify the aesthetic judgment of medicinal chemists \cite{bickerton2012quantifying},
and
 the synthetic accessibility score (SAS), 
 	which estimates the difficulty in synthesis for a given compound \cite{ertl2009estimation}.
We use the ZINC250K dataset, as extracted by \cite{gomez2016automatic} from the ZINC dataset \cite{irwin2012zinc}.
Given a molecule $C$, denote the logP, QED, and SAS as the components of a vector 
$\phi(C)=(\phi_\text{logP},\phi_\text{QED},\phi_\text{SAS})$.
We construct a function $f_P : \mathcal{L} \rightarrow \real^3$ that estimates $\phi(C)$, 
implemented as a neural network with two hidden layers (sizes 120 and 60),
	trained using MSE loss, and regularized with dropout ($p=0.5$) and weight decay.

The prediction results are shown as the distribution of relative errors (i.e.\ $ | \phi_i(C) - f_P(C)_i | / \text{mean}(\phi_i) $) on a held-out test set of size 20000  in Fig.\ \ref{chemprop} (top row). 
The mean and SE for logP, SAS, and QED are
$ 0.26 \pm 0.002 $,
$ 0.1 \pm 0.0006 $,
and 
$ 0.14 \pm 0.001 $ (PC: $0.8$, $0.87$, and $0.37$). 
This suggests that the estimated values of the various chemical drug-likeness properties are within a reasonable range of their true values.

\section{Molecular Optimization}



\tikzset{%
	block/.style    = {draw, thick, rectangle, minimum height = 1.9em,
		minimum width = 1.9em},
	fm/.style      = {draw, circle, node distance = 1.7cm}, 
	fm2/.style      = {draw, circle, node distance = 1.7cm}, 
	input/.style    = {coordinate}, 
	output/.style   = {coordinate}, 
	fitting node/.style={
		inner sep=0pt,
		fill=none,
		draw=none,
		reset transform,
		fit={(\pgf@pathminx,\pgf@pathminy) (\pgf@pathmaxx,\pgf@pathmaxy)}
	},
	reset transform/.code={\pgftransformreset}
}
\begin{figure}
	\begin{minipage}{0.6\textwidth}
	\centering
	\begin{tikzpicture}[auto, thick, node distance=1.9cm, >=triangle 45]
	\draw 
	node at (0,0)[fm,right=0mm,name=p,fill=green!30]{$P$}
	node [block, right of=p,right=-8mm,xshift=-1.7mm] (fs) {$f_S$}
	node [fm,right of=fs,fill=red!30,right=-1mm] (c) {$C$}
	node [below of = c, node distance=2.1cm] (k2) {}
	node [block,below right of=c,right=1mm] (fp) {$\nabla f_P$}
	node [block,right of=c] (ft) {$ \nabla f_T $}
	node [block,above right of=c,right=1mm] (fb) {$\nabla f_B$}
	node [block, right of=ft] (en) {$\nabla \mathcal{E}_P(C)$}
	node [below of = en, node distance=2.1cm] (k1) {};
	\draw[->](p) -- (fs);
	\draw[->](fs) -- (c) node [midway, above,xshift=-1.17mm] {$t=0$};
	\draw[->](c) -- (ft);		
	\draw[->](c) -- (fb);
	\draw[->](c) -- (fp);
	\draw[->](fb) -- (en);
	\draw[->](ft) -- (en);
	\draw[->](fp) -- (en);
	\draw[thick] (c.south) |- (k2) -- (k1.center) -- (en);
	\draw[->](k2) -- (c);
	\draw(k1) -- (k2) node [midway, below] {$t>0$};
	\draw[->](p) to [out=75,in=180] (fb);
	\end{tikzpicture} %
	\end{minipage}\hfill
	\begin{minipage}{0.4\textwidth}
	\begin{algorithm}[H]
		\caption{Molecular Optimization}\label{euclid}
		\begin{algorithmic}[1]
			\Procedure{LatentOpt}{$P$, $T$, $\eta$}
			\State $ C_0 = f_S(P) $
			\For{$t= 1$ to $T$}
			\State $ v \leftarrow \nabla \mathcal{E}_P(C_{t-1}) $
			\State $ C_t \leftarrow \text{ADAM}(C_{t-1}, \eta, v) $  
			\EndFor
			\State \textbf{return} $C_T$
			\EndProcedure 
		\end{algorithmic}
	\end{algorithm}
	\end{minipage}
	\caption{Latent space molecular optimization algorithm. The input protein target site $P$ is used to first construct and then iteratively improve a molecule $C$ via gradient descent on an energy $ \mathcal{E} $.}
	\label{schemata}
\end{figure}

Given the differentiable models above, it is straightforward to define the gradient descent procedure in the latent space (depicted in Fig.\ \ref{schemata}).
We need only to define the energy we are minimizing:
\begin{align}
	\mathcal{E}_P(C)    &= {E}_B(C,P) + {E}_P(C) \label{eq:energy}  \\
	 {E}_B(C,P)         &= \alpha_1 p_B(C,P) + \alpha_2 g_h( \widehat{B}_{\text{DSX}}(C,P) )\\
	{E}_P(C) 	    &= \gamma_1 g_q( \widehat{\phi}_\text{logP}(C) ) + \gamma_2 				
							\widehat{\phi}_\text{QED}(C) 
							+ \gamma_3 \widehat{\phi}_\text{SAS}(C)
							+ \gamma_4 f_T(C) 
\end{align}
for a fixed target input $P$ and variable latent chemical $C$,
where $ f_P = (\widehat{\phi}_\text{logP}, \widehat{\phi}_\text{QED}, \widehat{\phi}_\text{SAS}) $ and $f_B=(p_B, \widehat{B}_{\text{DSX}}) $.
The first term in equation \ref{eq:energy} acts as the targeting term, ensuring the resulting chemical has good predicted binding affinity to the target; the second term encourages the molecule to have pharmacologically desirable chemical properties.
We also view this latter intrinsic energy as a regularizer, preventing the optimization from exploiting anomalous behaviour of the affinity predictor. 
We chose $ \alpha = (-10, 1/200) $ and $\gamma = (-0.5, -1, 0.1, 1) $.

Two terms have additional transformations applied to them.
The first is an offset rectifier on the predicted DSX: $ g_h(x) = \max(-250, x) $, 
used to limit
the term's influence when it had started to reach the edge of biophysically plausible values (few real DSX values were below $-300$).
The second is a quadratic function on the logP: $ g_q(x) = -4x(x-5)/25 $, which encourages the logP to stay near the middle of $[0,5]$.
While sufficiently high logP is required for physiological transport of the drug, 
	a logP $> 5$ violates Lipinski's ``rule of five'' \cite{lipinski2001experimental}
	and a logP $> 3$ increases probability of toxicity \cite{waring2015analysis}; 
	as such, some literature suggest aiming for a logP within ${\sim}$1-3  \cite{waring2010lipophilicity}.

\begin{figure}
	\includegraphics[width=0.3231\textwidth,trim={16mm 16mm 16mm 16mm},clip]{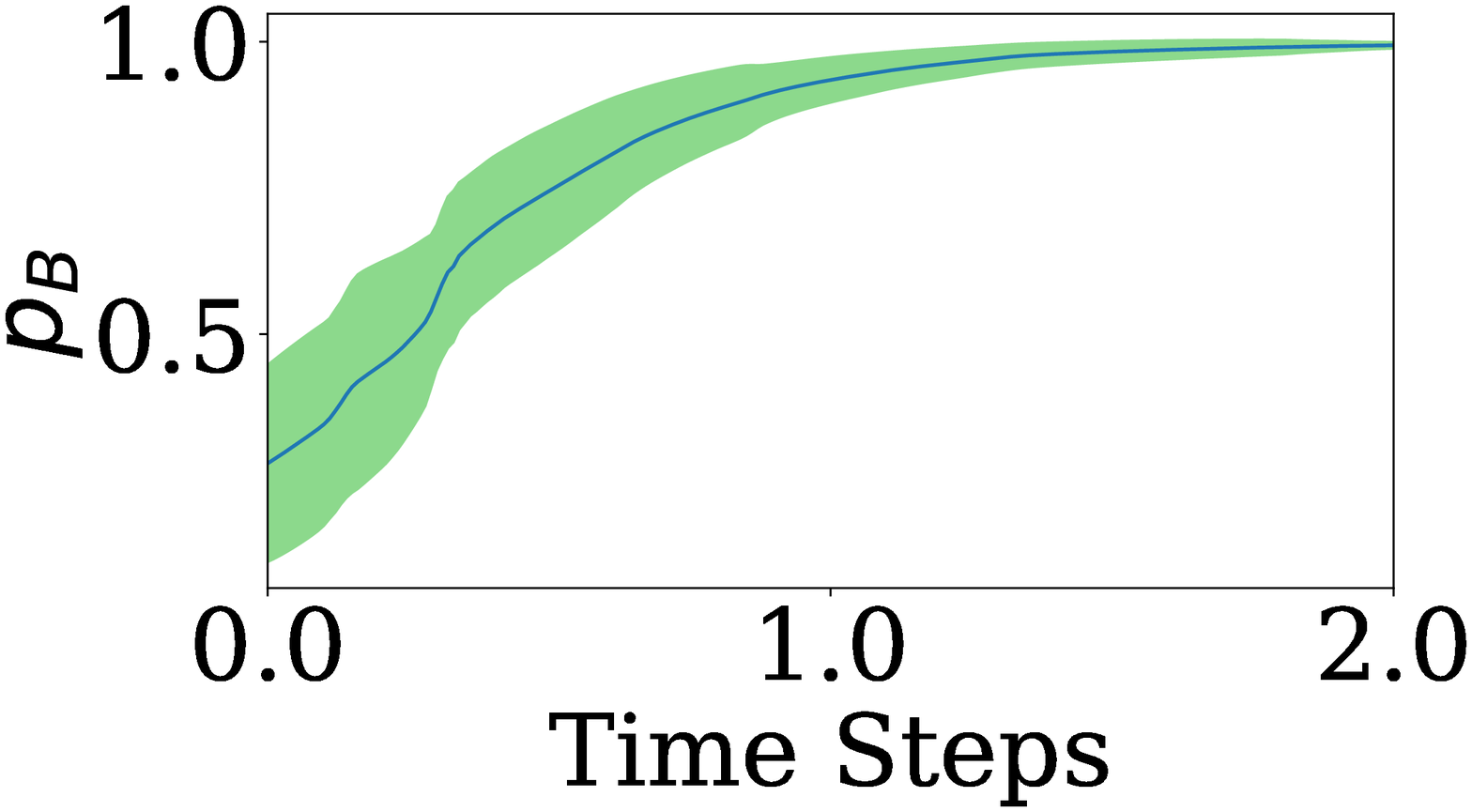}    \hfill
	\includegraphics[width=0.3231\textwidth,trim={16mm 16mm 16mm 16mm},clip]{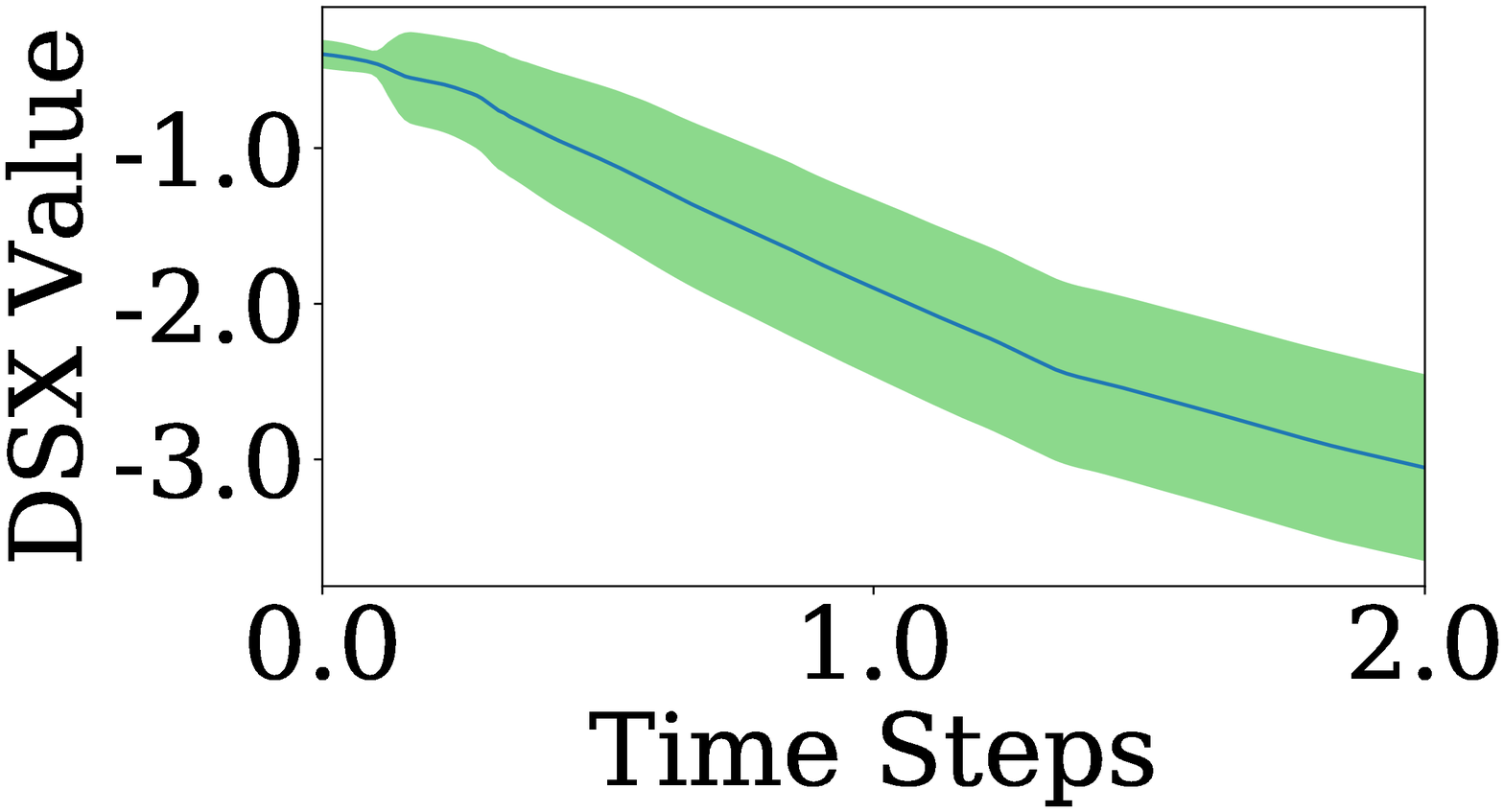}  \hfill
	\includegraphics[width=0.3231\textwidth,trim={16mm 16mm 16mm 16mm},clip]{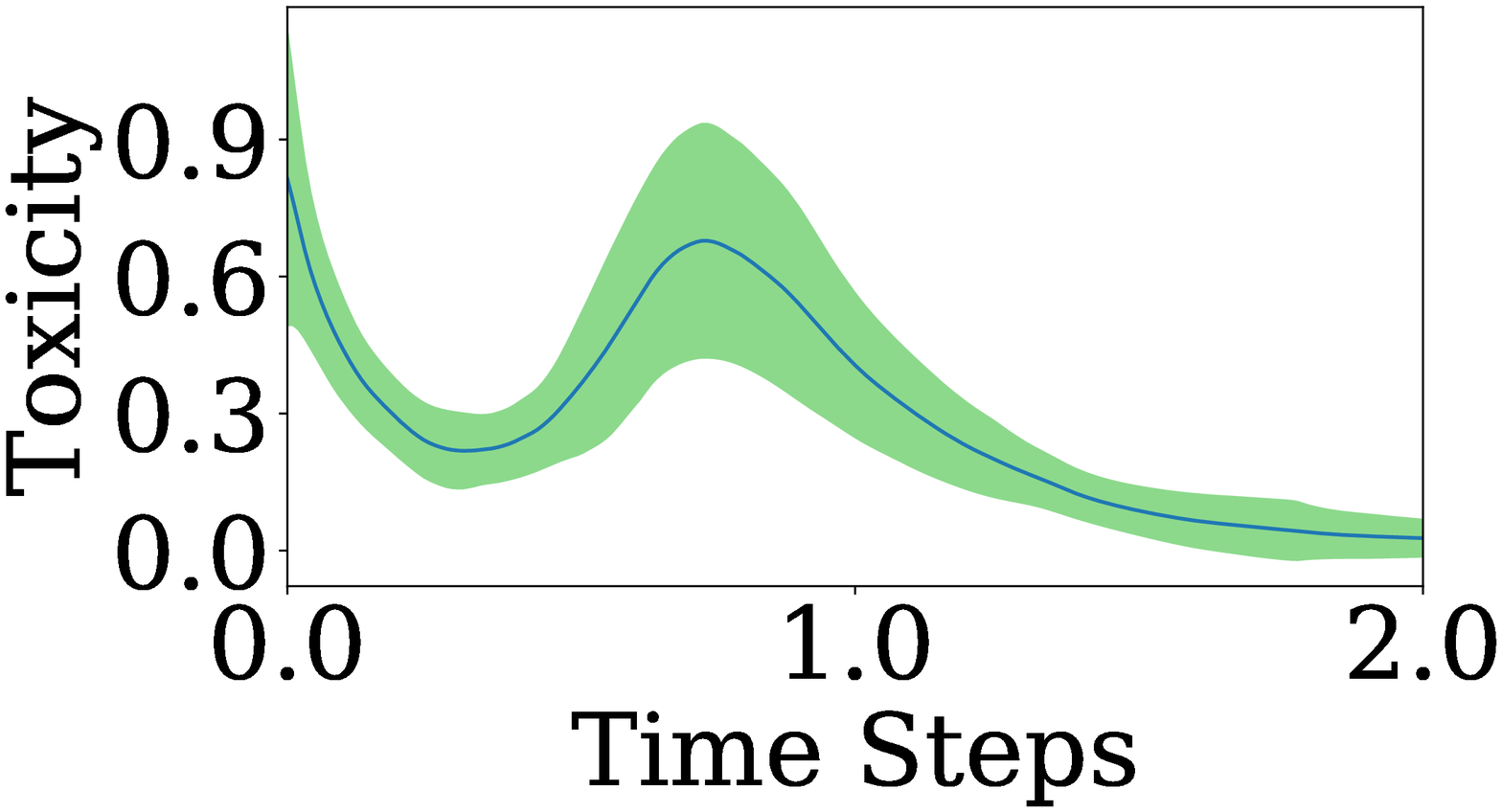}  \\
	\includegraphics[width=0.3231\textwidth,trim={16mm 16mm 16mm 16mm},clip]{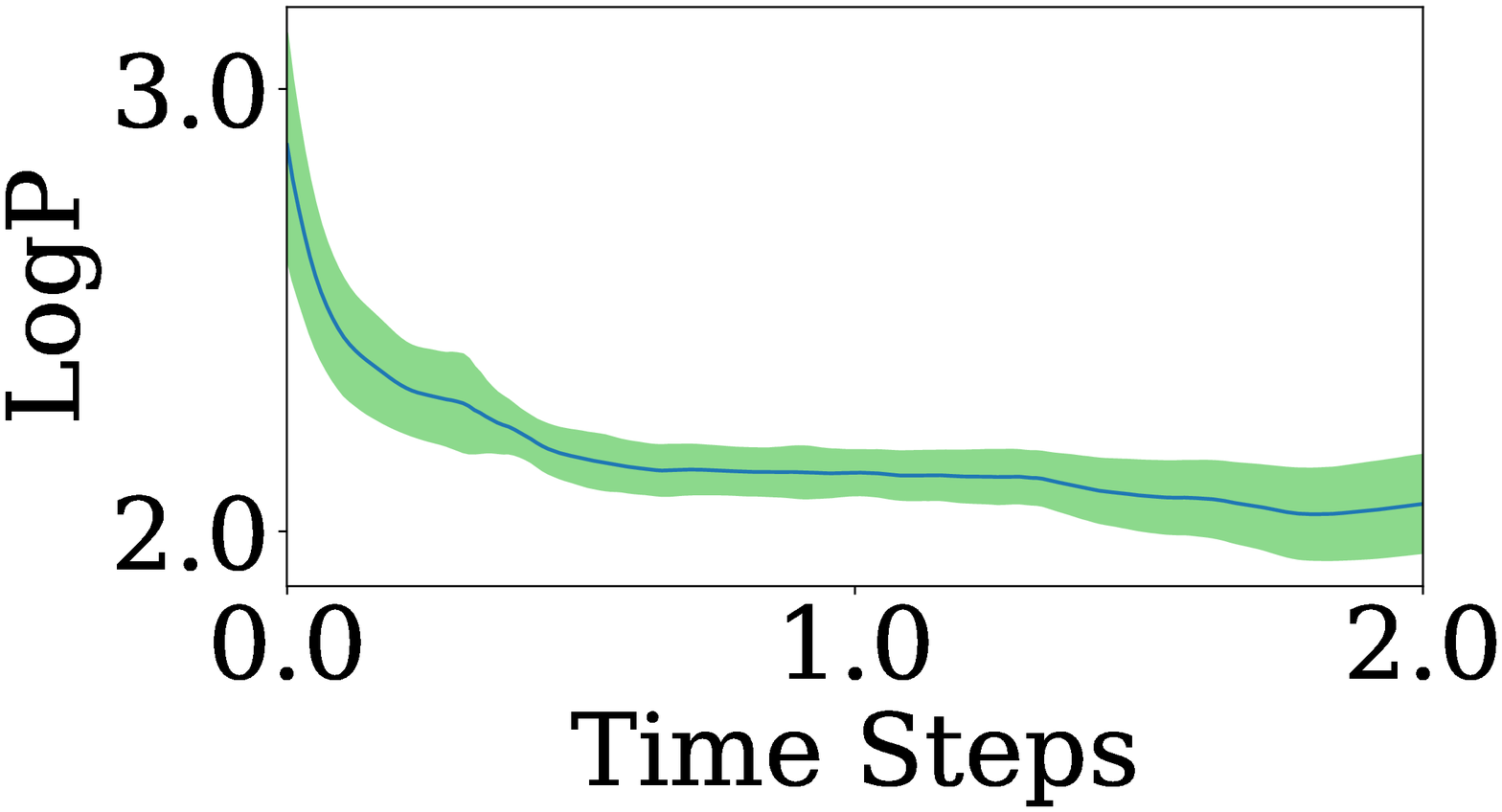} \hfill
	\includegraphics[width=0.3231\textwidth,trim={16mm 16mm 16mm 16mm},clip]{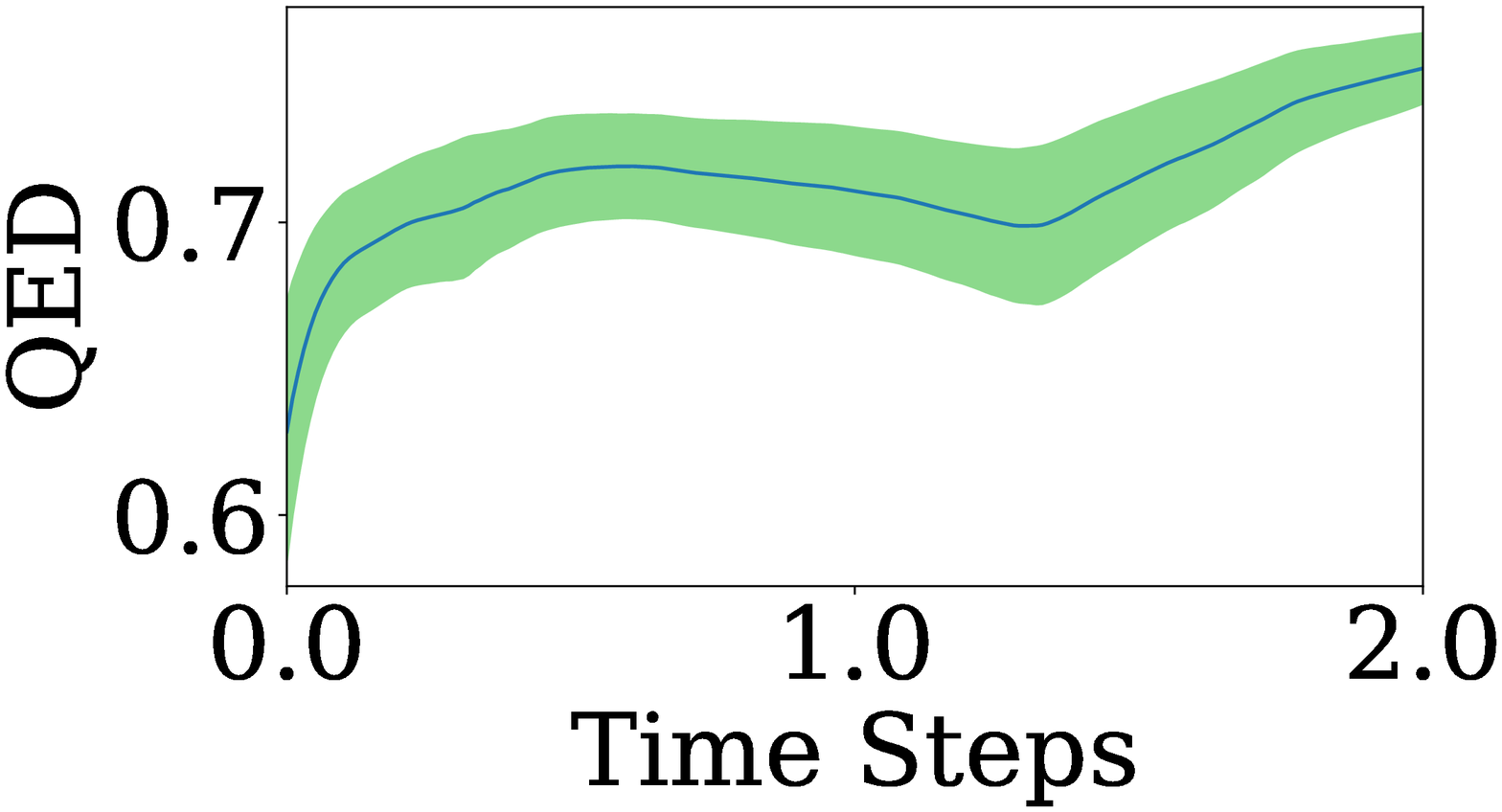}  \hfill
	\includegraphics[width=0.3231\textwidth,trim={16mm 16mm 16mm 16mm},clip]{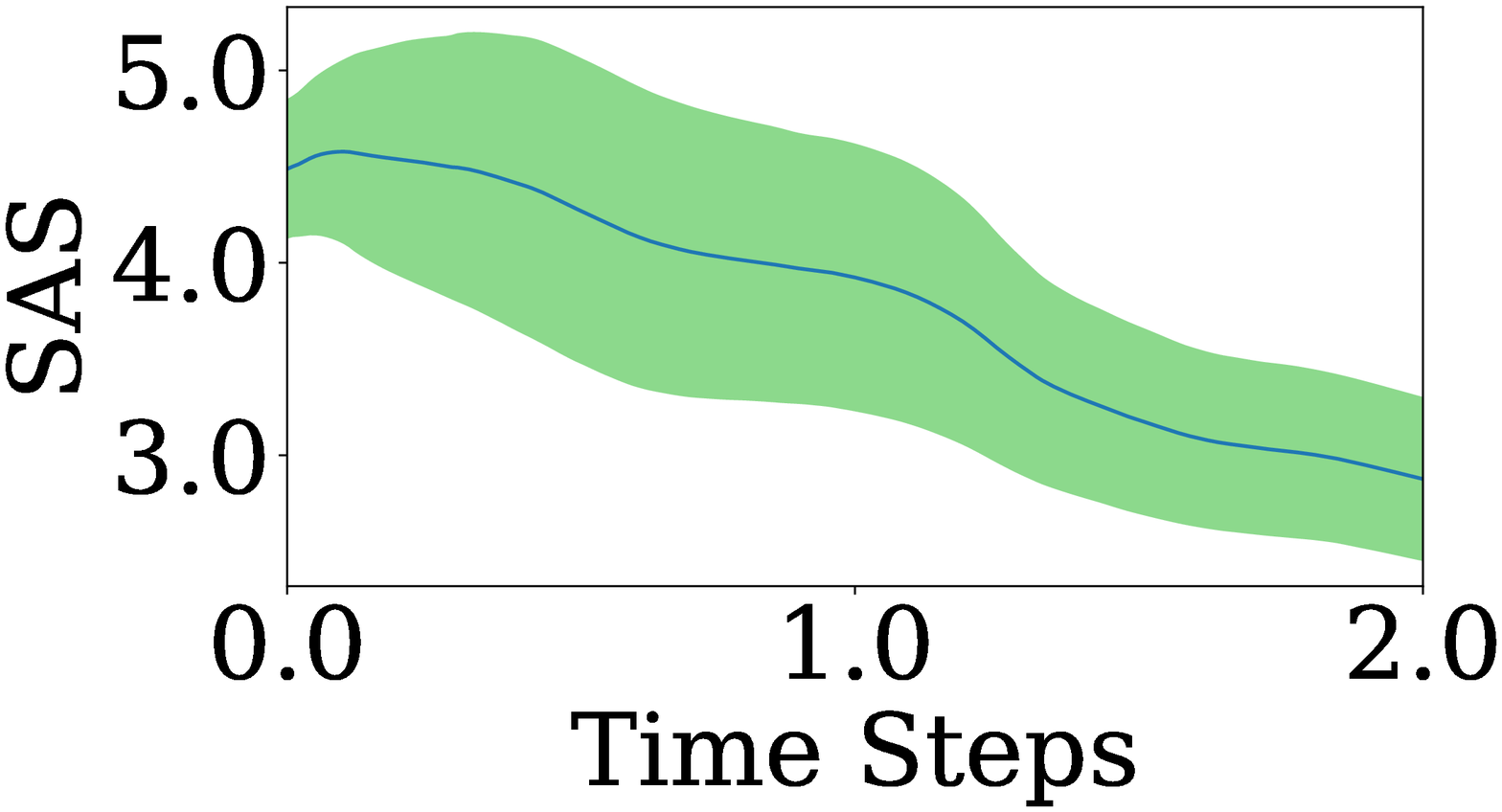} 
	\caption{
		Energy function component values during optimization.
		Blue line is average value, over the 1000 members of the test set; the green shaded area is the standard deviation (over PLCs) per time point.
		Time is in units of $10^4$ steps.
		Top: estimated binding probability, DSX (in hundreds), and toxicity.
		Bottom: estimated logP, QED, and SAS.
	} \label{energy}
\end{figure}

We tested our optimization algorithm on the target protein sites from the held-out PLC test set of size 1000 utilized above (i.e.\ they were unknown to any of the models).
The resulting components of the energy function over time are shown in Fig.\ \ref{energy}.
It is clear that the predicted binding probability and DSX are indicating strong estimated affinity for the majority of input targets by the end of the optimization. All of the intrinsic properties are also improved: logP stays near 2, while QED steadily increases, and both toxicity and SAS decrease.

\begin{figure}
	\centering 
	\begin{tabular}{>{\centering\bfseries}m{0.35\textwidth}  |  >{\centering\bfseries}m{0.35\textwidth} | >{\centering\arraybackslash}m{0.2\textwidth}}  
		\includegraphics[width=0.13\textwidth]{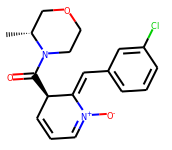} 
		\includegraphics[width=0.13\textwidth]{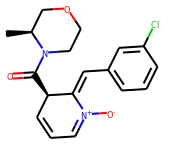}
		&
		\includegraphics[width=0.16\textwidth]{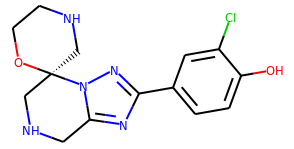} 
		\includegraphics[width=0.16\textwidth]{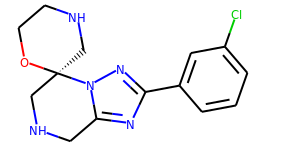}
		&
		\includegraphics[width=0.18\textwidth]{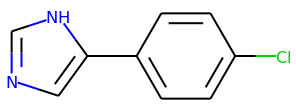} \\[-1mm]
		\includegraphics[width=0.13\textwidth]{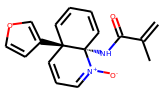}
		\includegraphics[width=0.13\textwidth]{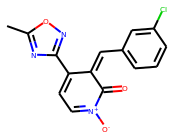}
		&
		\includegraphics[width=0.1406\textwidth]{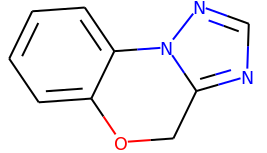} 
		\includegraphics[width=0.1406\textwidth]{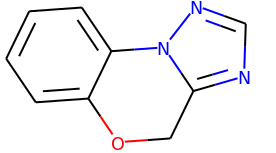}
		&
		\includegraphics[width=0.16\textwidth]{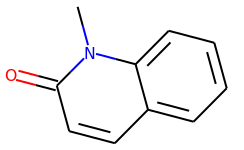}\\
		\includegraphics[width=0.145\textwidth]{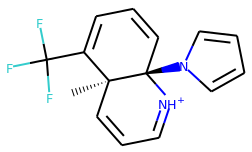} 
		\includegraphics[width=0.145\textwidth]{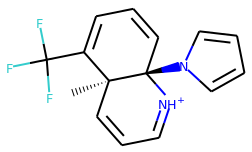}
		&
		\includegraphics[width=0.1451\textwidth]{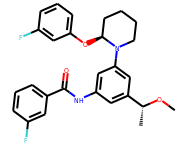} 
		\includegraphics[width=0.1451\textwidth]{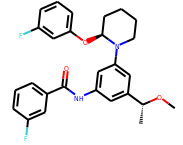}
		&
		\includegraphics[width=0.1758\textwidth]{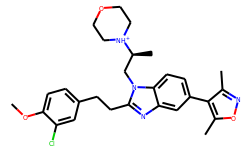}
	\end{tabular}
	\caption{Visualization of moieties found during optimization of target sites from held-out-test set. 
		Left: start of algorithm (output of $f_S$). 
		Middle: output of gradient descent after 20000 steps. 
		Right: known binder chemical. 
		Each row corresponds to a single input $P$.
		Two latent space samples are shown for the start and end chemicals.}
	\label{qual}
\end{figure}

Next, we qualitatively visualize some results. 
While we cannot say whether the chemicals output by the latent space optimization are good binders in reality or not, we can check if they have chemical moieties in common with the known binding ligand (which neither the algorithm nor the affinity model had access to), which would suggest the model learned a structural pattern in the protein site and associated latent molecule data that generalized to unseen graphs.
See Fig.\ \ref{qual} for examples of this phenomenon occurring.
For instance, in the first row of the figure, the known binder is close to being a subgraph of the output chemicals, 
while in the second and third row, the resulting chemical after optimization is qualitatively more similar to the binding ligand.

\begin{figure}
	  \begin{minipage}[c]{0.655\textwidth}
	  	\vspace{0pt}
		\includegraphics[height=0.29\linewidth]{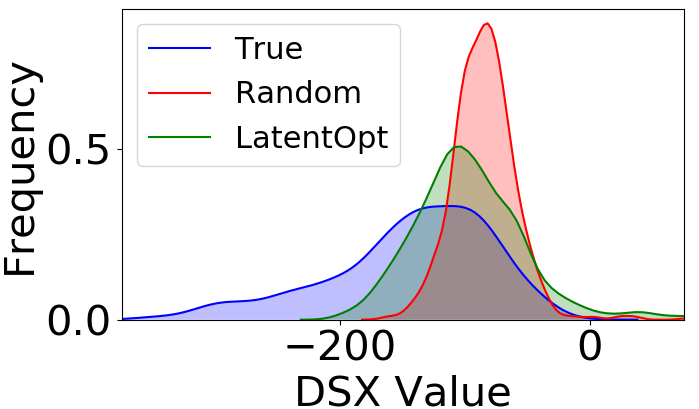}%
		\includegraphics[height=0.29\linewidth]{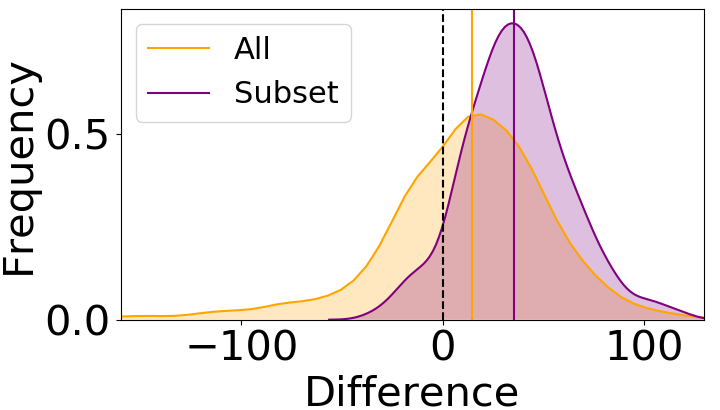}%
	\end{minipage}\hfill
	\begin{minipage}[c]{0.329\textwidth}
		\vspace{0pt}
		\caption{
			Docking validation results. 
			Left: DSX scores $D_T$ (blue), $D_R$ (red), and $D_C$ (green). 
			Right: DSX difference histograms of $\Delta$ (orange) and ${\Delta}_S$ (purple); lines at medians.
		} \label{docking}
	\end{minipage}
\end{figure}

Finally, we discuss quantitative evaluation of our optimization algorithm (see Fig.\ \ref{docking}).
This was done by docking generated ligands to their target proteins, and then estimating the DSX score of the bound complex; we compare the results to the DSX scores that result from generating random chemicals and docking one to each protein target.
We used Open Babel \cite{o2011open} for preprocessing and rDock \cite{ruiz2014rdock} for protein-ligand docking. 
In detail, for each target protein in the test set $P_i$, we generate two chemicals: $C_i$, the output of our optimization algorithm on $P_i$, and $R_i$, a ligand drawn from the latent prior of the JTVAE.
Using rDock, we then dock each of $R_i$ and $C_i$ to $P_i$ separately, obtaining two new bound complexes, from which we can estimate DSX scores $D_r(i)$ and $D_c(i)$.
For docking, since each $P_i$ has a known binder, the ``reference ligand'' approach to cavity generation is used, which restricts the input chemicals to bind to the same site as the known ligand; each docking is run 5 times, and we take the conformation with minimal DSX. 
Denote the ordered sets of DSX scores of the true, random, and latent binders as 
$ D_T $, $ D_R $, and $ D_C $ respectively; we also consider the set of differences 
$\Delta = \{ D_r(i) - D_c(i) \;\forall\; i \}$, so that a positive difference indicates that the optimized chemical has a better DSX score than its random counterpart.
Out of 1000 PLCs in the test set, we obtained 961 pairs of docked complexes.
In 631 of these (65.7\%), the optimized DSX was lower than its random counterpart;
further, while only 246 complexes with random ligands had a DSX less than $-100$, there were 477 optimized ligands with DSX score below that.
The DSX differences $\Delta$ had a median of 14.5, mean of 9.3, and standard deviation of 51.5.
Altogether, these results suggest the optimized ligands are enriched with better binders than one would expect to obtain by random sampling.

One issue is that, when the docking or scoring algorithm fails, we are essentially comparing 
two random ligands.
If we want to look only at the data for which all pipeline components appear to have produced reasonable results
(i.e.\ to see if it is better than random then), as well as eliminate physically implausible outliers,
we can restrict our attention to the case when the DSX is less than $-100$.
On this subset, 443 (92.9\%) of the optimized docked complexes have better DSX scores than their random counterparts.
The differences on this subset ${\Delta}_S = \{ D_r(i) - D_c(i) \,|\, D_c(i) < -100 \}$ had a median of 35.4, mean of 36.4, and standard deviation of 26.3.
Overall, while these results rely on simulated dockings, rather than empirical experiment, they still provide independent validation of the utility of our approach in generating targeted chemical ligands.

\section{Discussion}

We have proposed an algorithm for targeted therapeutic design:
	given an input site on a molecule of interest, we wish to generate an agent with satisfactory predicted binding affinity and chemical properties.
While both protein binding sites and ligands are inherently discrete and highly structured entities,
	the use of a graph convolutional signature extraction technique 
	and
	a deep generative latent variable encoder over chemical structures
	allows us to embed them into continuous vector spaces.
By learning differentiable models of target affinity and intrinsic molecular properties,
	we are able to utilize gradient-based optimization methods to perform targeted \textit{de novo} design.

In terms of future work, one could consider other 
scoring functions (e.g.\ based on learning \cite{ain2015machine}), 
models (e.g.\ manifold-theoretic geometric learning techniques \cite{bronstein2017geometric} or using adversarial generative models), 
data modalities (e.g.\ augmenting with biochemical data or QSAR models), and 
search regularizers (e.g.\ penalizing latent distance from the origin, to help ensure molecular validity).
Additional validation, e.g.\ via other docking methods or molecular dynamics, would also be useful, particularly given that docking can sometimes be unreliable \cite{ramirez2016reliable,ramirez2018reliable}.
Overall, our results suggest that this approach could be a promising avenue for future progress in computational drug design.


\subsubsection*{Acknowledgments}
We thank Michael Brudno and Izhar Wallach for suggestions and advice, as well as Xiaolei Liu for assistance in proof-reading and figure design.


\bibliographystyle{unsrt}
\bibliography{latdrug}

\clearpage
\includepdf{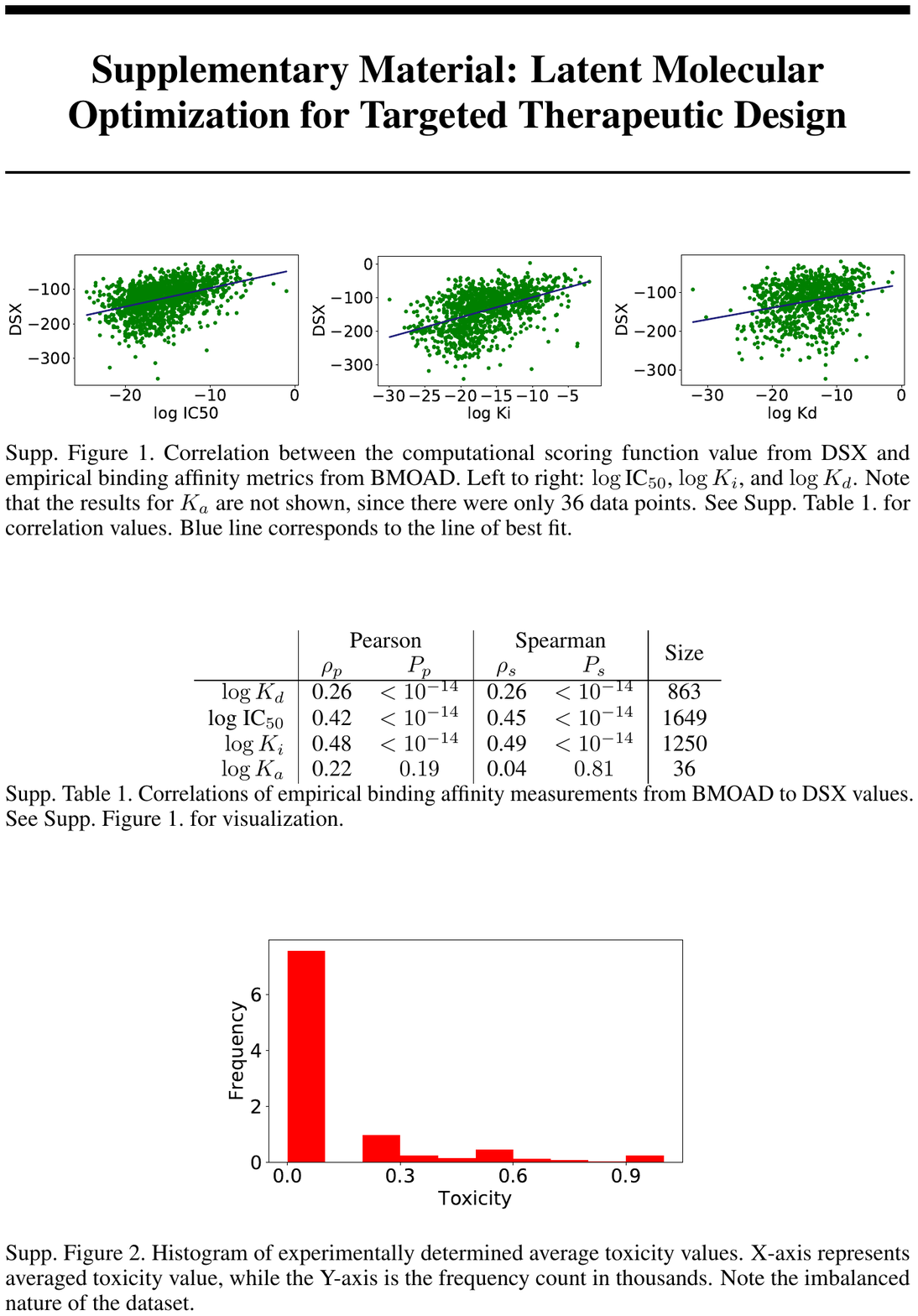}

\end{document}